\documentclass{article}

\usepackage[preprint]{neurips_2026}

\usepackage[utf8]{inputenc}
\usepackage[T1]{fontenc}
\usepackage{hyperref}
\usepackage{url}
\usepackage{booktabs}
\usepackage{amsfonts}
\usepackage{nicefrac}
\usepackage{microtype}
\usepackage{xcolor}
\usepackage{lipsum}
\usepackage{cleveref}
\usepackage{listings}
\lstdefinestyle{skillmd}{
  basicstyle=\ttfamily\scriptsize,
  breaklines=true,
  breakindent=0pt,
  postbreak=\mbox{$\hookrightarrow$\space},
  columns=fullflexible,
  keepspaces=true,
  frame=none,
  numbers=none,
  literate={—}{{---}}1 {→}{{$\rightarrow$}}1 {←}{{$\leftarrow$}}1 {–}{{--}}1
           {‘}{{`}}1 {’}{{'}}1 {“}{{``}}1 {”}{{''}}1
           {≤}{{$\leq$}}1 {≥}{{$\geq$}}1 {…}{{...}}1
           {×}{{$\times$}}1 {✓}{{$\checkmark$}}1
           {✅}{{$\checkmark$}}1 {❌}{{$\times$}}1
           {─}{{-}}1 {├}{{| -}}2 {└}{{` -}}2,
}

\crefname{part}{\S}{\S\S}
\crefname{chapter}{\S}{\S\S}
\crefname{section}{\S}{\S\S}
\crefname{subsection}{\S}{\S\S}
\crefname{subsubsection}{\S}{\S\S}
\crefname{figure}{fig.}{Fig.}
\Crefname{table}{Table}{Tables}
\Crefname{figure}{Fig.}{Fig.}

\ifdefined\hidetodos
  \newcommand{\todo}[1]{}
\else
  \newcommand{\todo}[1]{\textcolor{red}{\textbf{[TODO: #1]}}}
\fi
\newcommand{\ghrepo}[1]{\href{https://github.com/#1}{\texttt{#1}}}

\title{Skill Issue: Lessons from Optimizing Repository SKILLs for Coding Agents}

\author{%
  Mykhailo Kozyrev \\
  Technical University of Munich, JetBrains Research \\
  Munich, Germany \\
  \texttt{mykhailo.kozyrev@jetbrains.com} \\
  \And{}
  Andrei Kozyrev \\
  JetBrains Research \\
  Berlin, Germany \\
  \texttt{andrei.kozyrev@jetbrains.com} \\
  \And{}
  Anton Podkopaev \\
  Constructor University Bremen, JetBrains Research \\
  Amsterdam, Netherlands \\
  \texttt{apodkopaev@constructor.university} \\
}

\begin{document}

\maketitle

\begin{abstract}
  Coding agents increasingly read repository knowledge from SKILLs --- plain \texttt{.md} files versioned alongside the code. Recent work synthesizes these files automatically, by optimizing the document against a benchmark. A bare repository comes with no benchmark, and the synthetic tasks prior work builds are small enough that a capable agent saturates them with no document at all. We mine harder tasks --- merged pull requests of the repository, reverted at a single frozen base commit; and score a candidate document by whether the same agent does better with it than without it. On three Kotlin repositories, the documents GEPA finds raise this score by $4.9$pp on average, and the ones SkillOpt finds leave it where it started, $0.1$pp above the seed. The GEPA gain matches what prior work reports with the same optimizer, and at the dataset size a single repository supplies it cannot be separated from the agent's run-to-run variance; settling that would take more tasks than one repository's history yields. The documents themselves read better than the score: a maintainer of one repository found in them knowledge one only gets by working in the project.
\end{abstract}

\section{Introduction}\label{sec:intro}

Large language models are increasingly deployed as coding agents that operate with tools, domain context, and verifiers rather than as single prompts. As task complexity grows, the focus shifts from optimizing model weights to managing an agent's memory across tasks. Several attempts have been made to find the right representation for this memory~\citep{ace, reasoningbank, flex, skillweaver}. These representations, however, are not self-contained artifacts: reading the memory at inference time requires a standing service outside the repository. Retrieval runs against an embedding model and a vector store on every task, and, when the memory keeps self-evolving, an LLM-as-a-judge labels each new trajectory while an update step folds it back into the store~\citep{reasoningbank}. This separate infrastructure makes such methods hard to integrate into a developer's day-to-day workflow.

A growing body of work instead represents the distilled knowledge as reusable SKILLs\footnote{Agent's SKILLs: \url{https://agentskills.io}} --- plain \texttt{.md} files that live in the repository alongside the code. Being native artifacts, they can be versioned, reviewed in a pull request, and merged like any other source file, without a separate memory system to maintain. This lightweight approach is gaining traction rapidly: the \texttt{AGENTS.md} format is already used by over 60{,}000 open-source projects~\citep{agentsmd}, and the SKILL specification is supported by more than 45 agent tools across competing vendors~\citep{agentskills}.

While such SKILLs are often written by hand, a line of work seeks to synthesize them automatically from an agent's own runs. These works follow a common pattern: the agent's goal is framed as solving a particular task, and success is measured as a pass-rate on a given dataset~\citep{voyager, skillopt}. An optimization loop reasons over the agent's trajectories, proposes edits to the optimization target (the SKILL), and evaluates each edit by the agent's updated performance. We study the repository-specific case of this pattern: the SKILL describes a single code repository, and its job is to help a coding agent solve everyday tasks in it.

This case is underexplored, and the reason is the dataset the pattern presumes: a repository does not come with one, and no metric cleanly captures the \textit{``performance''} of an agent on it. Several works attempt to proxy it: by automatically constructing isolated test environments for agents operating in the repository~\citep{tessl2026skillopt}, or by mining SWE-bench~\citep{swebench}-style datasets constructed from the repository~\citep{gskill}. However, those approaches remain early-stage, and existing evaluations use synthetic tasks rather than real changes mined from the repository's history.

How much such a synthesized SKILL actually helps is, moreover, hard to read off the reported numbers. Gskill~\citep{gskill, gepablog}, the latter of these, lifts held-out resolve rate from $55\%$ to $82\%$ on one repository and from $24\%$ to $93\%$ on another, but measured against a reduced mini-swe-agent~\citep{minisweagent} driven by a small model, \texttt{gpt-5-mini}. On their strongest configuration, a Claude Code harness running Sonnet 4.5, the same benchmark leaves almost nothing to move: the agent reaches $100\%$ on the first repository with no skill at all, and $94.8\% \rightarrow 100\%$ on the second. At that resolution a binary pass rate is the wrong instrument, because it records only whether a task passed, not whether it passed for the right reason\@: \citet{agentlens} report that $10.7\%$ of \emph{passing} agent trajectories reach that verdict through blind retries or unverified edits rather than a principled solution. A gain of a few points is thus of the same magnitude as the share of passes the metric itself mislabels. The likely cause is the instrument rather than the method; the tasks come from SWE-smith~\citep{swesmith}, which injects defects into working code and so yields small, localized fixes rather than the changes a developer would make; we quantify this on their released data in \cref{sec:related}.

To address this, we change the instrument in two ways. First, we mine harder tasks --- real merged pull requests from a repository no benchmark covers, each specified by the tests the pull request itself adds or, when it adds none, by the repository's existing tests that fail once the change is reverted. The same agent that saturates the synthetic benchmark\footnote{The same harness, Claude Code, one model version later: gskill's transfer experiment runs Sonnet 4.5, while we run Sonnet 4.6.} resolves roughly $53\%$ of these with no SKILL, which leaves the headroom a skill would have to explain. Second, we drop the transfer assumption and optimize directly against the agent we intend to ship, rather than against a cheap proxy.

Concretely, we treat SKILL generation as an optimization of a single \texttt{.md} document against a metric mined from the target repository itself. The setup has two parts: a verifier that scores a SKILL variant, and a proposer that edits the SKILL\@. As proposers we compare GEPA~\citep{gepa}, a framework for optimizing any system with textual parameters against any evaluation metric, and SkillOpt~\citep{skillopt}, a strategy for self-evolving agent skills.

The verifier replays merged pull requests of the target repository. Each task reverse-applies one merged change at a base commit shared by every task, reintroducing the problem the change solved, and the tests that start failing become its specification. A Claude agent then attempts the task. The agent is the same on every run, with the same model, tools and limits, and only the SKILL varies. Rather than measuring a variant's pass rate, we compare each of its rollouts with the seed SKILL's rollout on the same task, so a variant gains nothing from tasks the agent already solves without it (\cref{sec:method}). Mirroring real pull requests is SWE-smith's~\citep{swesmith} own strategy, and their ablations rank it the most valuable one; we apply it outside Python and report the validity defects we ran into (\cref{sec:experiments}).

This work focuses on Kotlin repositories, following our internal interest in JVM tooling. We build such datasets for three of them: \ghrepo{kotest/kotest}, \ghrepo{ktorio/ktor} and \ghrepo{JetBrains/koog}~\citep{kotest,ktor,koog}, and optimize a SKILL for each of them with both proposers. The SKILLs found by GEPA raise the mean score by $4.9$pp and those found by SkillOpt by $0.1$pp. Neither gain is statistically significant on splits of this size, $20$ to $26$ held-out tasks per repository, and GEPA's is of the same order as what gskill reports with the same optimizer on its strongest configuration, and as the share of passes a binary verdict mislabels. On the metric we thus reproduce prior work's picture. We add what the number lacks: a bound on what a split this size can detect, and a maintainer's reading of the documents.

Looking at the runs behind the gain, we attribute the difficulty to the measurement rather than to the optimizers: the variance of the score compounds the run-to-run variance of the agent, so a difference of this size sits inside the spread the agent produces on its own, and the reflector is asked to choose between candidates separated by less than that. Converging under such a signal would, in practice, take more tasks than a repository supplies. For example, mining and validation leave $119$ tasks out of \texttt{koog}'s $660$ merged pull requests. Scoring one candidate on one task is a full agent rollout, \$$0.84$ on average in our runs, so a $200$-attempt optimization run costs hundreds of dollars (see \cref{tab:results}); the repository's history and the budget both run out before the split is large enough.

Inspecting the SKILLs themselves gives a more favourable picture than the score does, and it is also what exposed a defect in our own setup. Our first configuration followed SWE-bench and placed each task at its own parent commit; the optimizer then reflected over a repository spread across months of history and wrote claims that were false at the version the SKILL would ship against: module paths, build invocations, APIs that had since moved. Mining in the reverse direction at a single frozen base removes this by construction (see \cref{sec:method}), and only after this change did the optimized SKILLs consistently describe the state of the repository the agent runs against.

To assess the artifacts independently of the score, we asked a maintainer of one of the studied repositories, \texttt{koog}, to read the SKILLs both optimizers produced, and to review the agent's solutions to two open issues in the repository under three setups: no SKILL, the GEPA SKILL, and the SkillOpt SKILL (\cref{sec:exp-devs}). In short: both documents contain knowledge one only gets by working in the project, and both pad it with general advice the agent does not need. On the live issues, with either document the agent finished in under half the wall-clock time and at lower cost.

This work makes four contributions.\ (i) A pipeline that mines an optimization dataset from a repository's own merged pull requests, applied to JVM repositories at a single frozen base commit, together with the limits of what it can produce (\cref{sec:limitations}).\ (ii) An evaluation of two SKILL optimizers, GEPA and SkillOpt, over these datasets for three repositories, under an objective that scores each rollout against the seed SKILL's.\ (iii) An examination of the SKILLs themselves, including their review by a \texttt{koog} maintainer, showing that at this dataset size the pass-rate delta cannot carry the claim on its own.\ (iv) Numbers on what such mining yields: how many tasks survive, and what a single rollout costs.

The remainder of the paper is organized as follows. \Cref{sec:prelim} reviews the two proposers. \Cref{sec:method} defines reverse-PR mining, pairing, and the scores we rejected. \Cref{sec:experiments} answers four research questions on kotest, ktor, and koog: what the mining yields, what the documents gain, whether the gain is separable from the agent's variance, and how a maintainer reads them. \Cref{sec:related} situates the work against prompt optimization, repository context files, and gskill. \Cref{sec:limitations} states where the pipeline does not apply to a new repository.

\section{Preliminaries}\label{sec:prelim}

This section describes the two optimizers we instrument, GEPA and SkillOpt. Both were chosen because they optimize a natural-language artifact by reflecting over the agent's own rollouts, and because neither requires infrastructure beyond the repository at inference time; \cref{sec:related} places them among the alternatives.

\paragraph{GEPA.} GEPA~\citep{gepa, dspy} is a prompt optimizer. It treats the agent's prompt as the candidate being optimized and a strong reflection LM as the mutation operator. On each iteration, GEPA selects a candidate from the current Pareto frontier~\citep{pareto}, samples a small minibatch of training tasks, and runs the selected candidate on that minibatch to collect scores, rollout transcripts, and textual feedback. The reflection LM then uses this evidence to propose a rewritten candidate. If the rewritten candidate improves over its parent on the minibatch, GEPA adds it to the candidate pool and evaluates it on a larger selection set, which is used to maintain the Pareto frontier and to choose the final candidate.\

\paragraph{SkillOpt.} SkillOpt~\citep{skillopt} frames skill learning as optimization over an external natural-language state: the skill is a single \texttt{.md} document that a separate optimizer model edits while the target agent stays frozen. Each step runs the current skill on a batch of training tasks, and the optimizer reflects over minibatches of successes and failures to propose bounded \emph{add}/\emph{delete}/\emph{replace} edits. The number of edits applied per step is capped by an edit budget --- a textual analogue of a learning rate, decayed on a schedule --- and the merged edits are ranked by expected utility before the top ones are applied. A candidate skill is accepted only if it strictly improves a held-out selection score; otherwise it is rejected, and its edits enter an epoch-local rejected-edit buffer that keeps later proposals from repeating them. An epoch-wise slow/meta update behaves like momentum, folding durable lessons from adjacent epochs into a protected field of the document.

Both are reflection-driven optimizers over a natural-language artifact, but GEPA rewrites the whole instruction freely from a pool sampled off a Pareto frontier, while SkillOpt applies bounded, budgeted edits to one evolving document behind a strict improvement gate. We instrument both against the same validation score, to separate the effect of the metric from the effect of the optimizer.

\section{Methodology}\label{sec:method}

A SKILL, in this paper, is a single \texttt{.md} file that the harness loads into the agent's context before each task, and it is the only parameter we optimize: the coding model and the agent's topology stay fixed. Since almost no repository ships a benchmark, optimizing that file requires tasks the agent can attempt and a score over its runs; the proposers that read those runs and edit the document are GEPA and SkillOpt (\cref{sec:prelim}). This section defines the tasks, the score, and the evaluation procedure both proposers share.

Evaluating a candidate means running the same coding agent on a task with that document loaded. The agent works in an isolated copy of the repository with the project's history hidden, so it cannot read the original change and paste it back. The first candidate carries no advice; every later one is an edit of it. A run returns the agent's attempt, the edits it made, and how the tests then behaved.

\subsection{Task collection}\label{sec:tasks}

We target Kotlin, which SWE-smith~\citep{swesmith} does not support: its entity extraction runs through the Python \texttt{ast} module and its harness assumes \texttt{pytest}. We therefore reimplement its highest-ranked strategy, PR mirroring, for the JVM\@. We stream a repository's merged pull requests and split each diff by path into an implementation part and a test part over \texttt{.kt} and \texttt{.kts} files.

A merged change can be turned into a task in two directions. The forward direction is the SWE-bench setup: check out the commit the change was written against, apply its test patch, and ask the agent for the implementation. It yields the most tasks, since every merged change is usable at its own parent commit, but it anchors each task to a different, historical version of the repository, and two problems occur. Where pull requests depend on one another, the test patch does not compile at the parent commit at all, because the sibling changes it relies on had not merged yet. And a SKILL optimized over such a distribution describes the repository as it was across months of history rather than as it is: the SKILLs we obtained this way asserted module layouts and APIs that no longer exist, and the agent followed that advice into failures. \Cref{app:skill-outdated} shows one such document: its opening lines instruct the agent to identify which of \textit{``two distinct repository shapes''} it is working in --- one repository, seen at different points of its history.

We therefore mine in the reverse direction, reverting the implementation \emph{at a single frozen base commit} shared by every task, so that whatever a SKILL learns describes the current repository. There are two trade-offs to this. The first is the number of tasks: a change is only usable while the code it touched is still present at the frozen base, so a small or heavily refactored repository runs out of candidates (more in \cref{sec:limitations}). The second is that a change rarely reverse-applies at a base it was not written against, so we revert in three tiers, cheapest first. \texttt{git apply -{}-reverse} handles the changes the base has not drifted away from. Added and deleted files are then undone structurally, by re-creating or removing them. Only modified files that survive neither tier, and that fall within a size gate, are passed to an LLM, which is shown the current file together with the forward diff and asked to reconstruct the pre-change source. The patch is re-derived from \texttt{git} afterwards rather than taken from the model's output, so every instance is a real patch against the real base. Tests already exist at the frozen base, so a change that shipped none of its own stays usable: reverting it can still make an existing test fail, and validation finds which one.

The LLM tier carries most of the yield --- only $25$ of \texttt{koog}'s $119$ graded tasks reverse-apply cleanly, $56$ of $100$ on \texttt{kotest} and $65$ of $131$ on \texttt{ktor} --- and introduces defects of its own that the validation gate cannot see; a static check catches them (\cref{app:reversion-defects}).

Launching an agent on a dataset item requires a task description. Since our dataset items are pull requests, a natural choice for the problem description is the corresponding issue. However, issues are not always available: many changes are opened without one, and repositories that track work in an external tracker leave nothing linked on GitHub. The pull request's own description is the obvious fallback, but it is written \emph{after} the work is completed and often leaks the solution to the agent. When no issue is linked we therefore synthesise the statement instead, rewriting the pull request description into issue-style text with a single LLM call at collection time, following SWE-smith's finding that model-written issue text performs comparably to real issue text. Each synthesised statement is then checked against the gold patch for leakage. The repository leaks the solution as well, so links into the pull request are scrubbed, and each container strips \texttt{.git} and re-initialises a single commit before the agent starts.

\paragraph{Validation and grading.} A mined task is only worth a rollout if reverting its implementation actually breaks the tests meant to grade it. A task that already passes at the reverted base gives every candidate a score of $1$, and one that already fails for unrelated reasons gives every candidate a $0$; in both cases the task discriminates nothing, and neither condition is visible from the rollout itself. We therefore validate every task before admitting it. Validation runs inside a Docker image built once per repository. Its build and test commands are first discovered by the agent. The suite is then run once at the unmodified base to record which tests pass there; then, per task, the gold patch is reverse-applied and the suite re-run. The tests that switch from passing to failing become that task's FAIL\_TO\_PASS set --- the mechanism that also recovers grading targets for changes that shipped no tests of their own. A task is rejected if no test switches, if the affected tests were already failing, if the patch will not reverse-apply, or if the reversion breaks so much of the suite that it no longer isolates the change; each accepted task additionally carries up to $30$ regression guards sampled from the tests that passed at the base.

\subsection{The objective}\label{sec:objective}

Scoring a rollout on its own measures the repository rather than the document: the agent already resolves many tasks with no advice at all, so every candidate collects whatever the split makes easy, and a SKILL that changes nothing about the agent still keeps that score. We therefore score a \emph{pair}. The seed SKILL is run once on every task, and each candidate rollout is compared against the seed's rollout on the same task, under the same model, tools, limits and container. The number then answers one question: whether the agent did better with the document than without it.

The two rollouts are compared in three steps, and the first step that separates them decides. A rollout that edits the test files, or that reports a success its own tests contradict, loses to one that does not. Failing that, the rollout satisfying every hidden FAIL\_TO\_PASS test wins. Only when neither step separates them do the remaining quantities break the tie, the fraction of tests passed, the honesty of the final claim, the size of the diff, the number of tool calls, blended within a bounded range, so a cheaper or tidier rollout can never outrank a correct one. The seed against itself is exactly $0.5$, which makes $0.5$ the number a candidate has to beat, and makes a regression on a single task visible as a score below it. 

Before pairing we tried absolute scores: the number of tasks the agent solved outright, with every hidden FAIL\_TO\_PASS test passing, the distance between its diff and the merged one, the price of the run, a judge's verdict on the trajectory, and a grade given to the document itself. Each answers a different question than ours, and the failures bracket the range. Counting solved tasks leaves almost every rollout at $0$, so the proposer sees no difference between candidates and ships the seed. Grading the document improves the grade while leaving the agent unchanged.\footnote{\Cref{app:rejected} reports each of these scores, the document it selected, and why we dropped it.}

The seed is an empty document; every number we report is the same agent measured with it and without it.

\section{Experiments}\label{sec:experiments}

This section asks four questions. \textbf{RQ1}: how many usable tasks does the procedure described in \cref{sec:tasks} recover from a repository's merged pull requests, and how difficult are they compared to the tasks SWE-smith~\citep{swesmith} mines? \textbf{RQ2}: do GEPA and SkillOpt find a document that helps the agent on held-out tasks? \textbf{RQ3}: can a gain of the size we measure be attributed to the document, rather than to the agent's own variance on the same tasks? \textbf{RQ4}: do the resulting documents hold knowledge a maintainer of the repository finds useful?

\subsection{Setup}\label{sec:exp-setup}

The coding agent is Claude Code, driven by Sonnet~4.6. We leave that model frozen: the only thing that changes between candidates is the \texttt{SKILL.md} loaded into context. Each attempt runs in a fresh Docker container, from an image built once per repository, on an isolated checkout at the frozen base with \texttt{.git} removed. The tools and the limits are the same for every candidate; only the document differs.

We score a candidate by pairing it with the stored empty-seed run of the same task (\cref{sec:objective}).\footnote{Skill is actually present, but contains nothing but a header} Each repository's tasks are split once into train, validation, and test. GEPA is allowed $200$ scored attempts on each repository. SkillOpt runs three epochs on the same splits, with the test split left unseen.

The repositories are Kotlin/JVM projects that SWE-smith does not cover and whose test suite a sandbox can run. To answer \textbf{RQ1} we report two properties of the resulting pools. First, what the procedure of \cref{sec:tasks} yields. Of $660$ of \texttt{koog}'s merged pull requests, $119$ became graded tasks;\footnote{These counts are the most recent merged pull requests we streamed, not the repositories' full histories; we stopped once the surviving pool cleared $100$ tasks.} \texttt{ktor} gives $131$ out of $452$, and \texttt{kotest} $100$, the last additionally cut by a seeded random sample to bound the run's wall clock. Roughly one merged pull request in five survives, and the loss is spread over the stages: patches that no longer reverse-apply at the frozen base, reversions that break no test, changes with no test module to grade them, and tasks whose own gold patch fails to score $1$. No single filter can be relaxed to make the pool much larger.

Second, how difficult those tasks are. The instances SWE-smith releases for the repositories gskill optimizes against change a median of $4$ and $7$ lines and are confined to a single file in $98$--$100\%$ of cases (\cref{sec:related}). Ours are larger: a median of $54$ lines across $3$ files on \texttt{koog}, $27$ across $2$ on \texttt{ktor}, and $16$ within one file on \texttt{kotest}, with only $23\%$, $50\%$ and $73\%$ of tasks respectively touching a single file. The repositories differ, so the contrast is between ways of building tasks, not between projects. With no SKILL loaded the agent resolves $40\%$ of \texttt{koog}'s tasks, $51\%$ of \texttt{kotest}'s and $66\%$ of \texttt{ktor}'s, $53\%$ pooled, where the same harness leaves almost nothing to resolve on SWE-smith's.

\begin{table}[t]
\centering
\caption{Final test-set score and optimization cost per repository. \emph{Spend} is the total agent API cost of the run.}
\label{tab:results}
\small
\begin{tabular}{llccccrr}
\toprule
repository & optimizer & $\Delta$ test-set & paired & spend & wall clock \\
\midrule
\texttt{JetBrains/koog} & GEPA     & $+1$ & 0.525 & \$368.82 & 4.7\,h \\
                           & SkillOpt & $+2$ & 0.546 & \$401.44 & 8.14\,h \\
\addlinespace
\texttt{kotest/kotest} & GEPA     & $+3$ & 0.554 & \$182.48 & 4.9\,h \\
                           & SkillOpt & $+1$ & 0.519 & \$263.56 & 13.3\,h \\
\addlinespace
\texttt{ktorio/ktor}  & GEPA     & $+4$ & 0.567 & \$312.87 & 13.0\,h \\
                           & SkillOpt & $-2$ & 0.437 & \$484.81 & 25.2\,h \\
\midrule
total                      &           &      &       & \$2{,}013.98 & 69.2\,h \\
\bottomrule
\end{tabular}
\end{table}

\subsection{The measured gain}\label{sec:exp-rq2}

\textbf{RQ2} is settled on the split neither proposer saw. GEPA's document scores $0.554$ on \texttt{kotest}, $0.567$ on \texttt{ktor} and $0.525$ on \texttt{koog}, against the seed's $0.5$ against itself; counted in resolved tasks that is $13\to16$ of $20$, $16\to20$ of $26$ and $14\to15$ of $23$. SkillOpt, on the same splits, scores $0.546$ on \texttt{koog}, $0.519$ on \texttt{kotest} and $0.437$ on \texttt{ktor}. On the same tasks and the same seed, GEPA's document wins on every repository, while SkillOpt's loss on \texttt{ktor} cancels its gains elsewhere.

\textbf{RQ3} asks whether that gain belongs to the document or to the agent rerolling. If the document does nothing, the two rollouts of a task are exchangeable and the candidate wins half of them, so a sign test over the tasks where they disagree tests exactly that; no run of either proposer clears $p=0.05$, the best being $p=0.29$. The more useful number is what such a split could have shown at all: at $20$--$26$ tasks the test rejects only when the document wins four of every five disagreements, and pooled over $69$ tasks two of three. Every effect reported in this line of work sits below that line: ours, and gskill's on its strongest configuration. The documents may still help; however, a pass rate over a hundred mined tasks is not sensitive enough to show it. \Cref{sec:exp-devs} therefore puts it to a maintainer instead.

\subsection{A maintainer's reading}\label{sec:exp-devs}

\textbf{RQ4} puts the two \texttt{koog} documents in front of a maintainer of that repository, section by section. Of the one GEPA produced (\cref{app:skill}), the parts they marked as worth having are the ones a reader cannot get without working in the project. On multiplatform source sets, Kotlin's mechanism for splitting a module's code between target platforms, \textit{``we had a lot of pain with them, agents often do not understand what is going on there''}; the rule that code must not reach across source-set boundaries is \textit{``all correct, good that it is included''}; the prescribed direction of dependencies between modules is \textit{``very important''}. Their request there is for more of the same: the multiplatform section \textit{``needs to be written in even more detail, but a developer has to do that, because the details have to be spelled out''}. The misses are as specific. The document never mentions \texttt{@Tool}, the annotation that turns a Kotlin function into a tool the agent can call and \textit{``the simplest way to define a tool''}, which they call \textit{``very strange''}; on \texttt{expect}/\texttt{actual}, Kotlin's way of declaring an API once and implementing it per platform, it states the rule but omits the pattern the team uses to avoid duplicating code, \textit{``it took us a while to work out how to avoid that''}; and on the least stable APIs it goes into detail the code will not honour, since \textit{``the API is less stable than that level of detail implies''}. One section, on checking claims about the public API against the repository's generated listings of it, they read as having missed its own point.

In the document SkillOpt produced (\cref{app:skill-skillopt}) the maintainer finds less to keep. One item is singled out as \textit{``great and important''}: that registering a new module is a three-file change, which is non-trivial in \texttt{koog} and nowhere written down, while the rest reads to them as \textit{``a lot of description of very general best practices''}, of the kind that \textit{``the agent should know by itself''} and that belongs \textit{``in a global skill, and only the repository-specific pieces kept in the repo''}. Several sections are valid and too small to matter: repository conventions, explicit API mode, multi-module build hygiene. This is one maintainer of one repository, and what we collected is an opinion of what the two documents contain.

We asked whether they would merge either document if it arrived as a pull request, the maintainer takes the GEPA one: they \textit{``would take the first as a draft and polish it''}, and \textit{``as a draft it is super''}; the SkillOpt one \textit{``needs more polishing''}.

The review closes with two issues that were open in the repository when the experiment was set up: \href{https://github.com/JetBrains/koog/issues/1275}{\#1275}, on reporting cached token counts, and \href{https://github.com/JetBrains/koog/issues/1354}{\#1354}, on attaching response headers to a thrown error. The agent solved each with no SKILL, with the GEPA document and with the SkillOpt one, and the maintainer read the three patches per issue in an anonymized diff viewer, not knowing which produced which. With either document the agent finishes faster and cheaper on both: end to end it takes $15$ and $20$ minutes with no SKILL against $6$ to $8$ minutes with one, at \$$6.06$ and \$$4.85$ against \$$2.44$ to \$$4.20$ (see table in \cref{app:issue-cost}; the patches are in \cref{app:issue-patches}). On \#1275 the three patches are \textit{``quite similar''}: the GEPA one is \textit{``notably more minimalist, which is good''}, the no-SKILL one \textit{``does a bit extra''} and leaves \textit{``strange comments in several places''}, the SkillOpt one \textit{``looks fine''}, and all three lack an integration test. On \#1354 the solutions are \textit{``overall ok''}: the GEPA one is \textit{``a bit tidier''} than the no-SKILL one, both of those updated the repository's generated API listings unprompted, and the SkillOpt one is \textit{``almost identical''} to the GEPA one.

\section{Related Work}\label{sec:related}

We generate a SKILL for a code repository by optimizing its text against a metric mined from that same repository. Prior work covers these components separately: optimizers that search over text, the repository-level file formats a SKILL is written in, and earlier attempts to synthesize such files automatically. We review each in turn.

\paragraph{Prompt and context optimization.} Optimizing an agent's textual context against a scored objective is by now a well-established setting. DSPy~\citep{dspy} provides a general framework for it, GEPA~\citep{gepa} evolves prompts by reflecting over rollouts (\cref{sec:prelim}), and systems such as ACE~\citep{ace} and ReasoningBank~\citep{reasoningbank} accumulate distilled experience across tasks and feed it back into subsequent runs. SKILL optimization is a close relative of this line: the target is still a natural-language artifact, and the optimizer is still a reflection loop over trajectories. What these approaches share, however, is a structural assumption --- a task distribution and a scalar metric are given in advance --- and a bare code repository supplies neither. Our setting therefore inherits their machinery but not their premise: constructing the metric becomes part of the problem rather than an input to it.

\paragraph{Repository-level context files.} A parallel line places the distilled knowledge in the repository itself, as plain \texttt{.md} files --- the \texttt{AGENTS.md} format~\citep{agentsmd} and the SKILL specification~\citep{agentskills}. We adopt this representation for the reasons given in \cref{sec:intro}; for an optimizer, it imposes an additional constraint: the output must remain a document a developer is willing to review and maintain.

Whether such files help at all is, however, unsettled. \citet{ctxbench} evaluate developer-committed and LLM-generated \texttt{AGENTS.md}/\texttt{CLAUDE.md} files across four agent and model pairs, on SWE-bench Lite and on a benchmark of tasks mined from repositories that already ship them. Files produced by the agents' own initialization commands do not improve the resolution rate --- $-0.5\%$ and $-2\%$ on average, neither significant --- while raising cost by $20\%$--$23\%$; developer-written files gain $2.4\%$, also not significant. Yet once the repository's remaining documentation is stripped away, the same generated files land slightly ahead ($+2.7\%$), which suggests that in their default form they largely restate the README\@.

\paragraph{Automatic skill synthesis.} Rather than authoring such artifacts by hand, several works synthesize them from an agent's own runs. Voyager~\citep{voyager} grows a library of executable skills from successful rollouts in an environment with a dense, automatic reward, and SkillWeaver~\citep{skillweaver} applies the same idea to web agents; in both cases the artifact is code rather than prose. FLEX~\citep{flex} does accumulate a natural-language experience library through reflection on successes and failures, but it is evaluated on mathematical and scientific reasoning, where a ground-truth answer is always available. SkillOpt~\citep{skillopt} makes the optimization over a natural-language document explicit (\cref{sec:prelim}). For code, the metric must first be defined: skill-optimizer~\citep{tessl2026skillopt} constructs isolated test environments for the agent, while gskill~\citep{gskill} generates task datasets over a target repository.

GEPA's authors apply their optimizer to this problem directly --- automatic SKILL synthesis for a single repository --- in gskill~\citep{gskill} and an accompanying blog post~\citep{gepablog}. It is the closest published system to ours; its held-out numbers and their saturation under a production agent are discussed in \cref{sec:intro}. The optimization target itself invites the scaffolding those numbers hide: gskill evolves the skill slot of its own agent template, whose instance prompt is $34$ characters long, while the stock mini-swe-agent~\citep{minisweagent} template shipped beside it is $2{,}369$ characters and carries the recommended workflow, the file-editing recipes, and the submission command. Part of what the optimizer recovers into the skill is therefore scaffolding the target agent does not need --- consistent with the published excerpt, whose advice (``run tests early, start broad with \texttt{go test ./...}'', ``avoid scratch \texttt{main.go} files in repo root'') is generic agent hygiene rather than repository knowledge~\citep{gepablog}; the authors themselves caution that some of these skills help ``SWE-smith style tasks (fixing issues)'' more than general coding practices. This reading comes from their released configuration and published excerpt, not from rerunning their optimization, and we make no claim about their reported rates.

The tasks themselves are the second half of the instrument: SWE-smith~\citep{swesmith} takes the repository's existing tests as the specification and writes issue text with a model; the authors note that the resulting instances are comparatively simple. SWE-smith's own ablations, though, rank \emph{PR mirroring} (\cref{sec:tasks}) as its most valuable strategy, and it is almost absent from the released data: none of the $351$ instances of the Go repository gskill uses are PR-mirrored, and $1$ of the $957$ Python ones\footnote{HuggingFace \texttt{SWE-bench/SWE-smith}, revision \texttt{ea6d717}.}. What remains is procedural and LLM mutation, and every patch on the Go repository is confined to a single file. A further $37.6\%$ of that repository's instances ship no problem statement at all; gskill's evaluation drops them, but its optimization path does not, so part of the signal it reflects over comes from tasks the agent was never told about.

\section{Limitations}\label{sec:limitations}

The procedure of \cref{sec:tasks} does not automatically produce a usable optimization set. Both proposers need three disjoint splits --- a train split the proposer edits against, a validation split that keeps or rejects a rewrite, and a test split neither sees --- so we run them on a repository only once mining and validation have left at least $100$ graded tasks. That floor is what makes three splits possible, not evidence that it suffices: it yields held-out splits of $20$ to $26$ tasks, and at that size a pass-rate comparison cannot separate a few-point gain from the agent's own variance (\cref{sec:exp-rq2}).

Two repositories we tried never reached it, each failing a different check. \ghrepo{JetBrains/tracy}~\citep{tracy} fails reverse-application: a package rename left none of its $213$ production files unchanged over three months, so almost none of its historical diffs still applied at the frozen base, and validation kept $5$ tasks out of $203$ merged pull requests. \ghrepo{http4k/http4k}~\citep{http4k} fails validation instead. Of $700$ listed pull requests, $150$ still reverse-applied, but reverse-application asks only whether the diff still applies, not whether reverting it makes any test fail in isolation, and fewer than $100$ of those survived the second check. \Cref{app:limitations-extended} reports what each check drops, and what a new repository has to supply to clear the bound.

\section{Conclusion}

Optimizing a SKILL for a repository requires a task distribution and a metric that the repository does not come with, so both have to be built. Prior work builds them by injecting defects into working code. The resulting tasks are small and mostly single-file, and a capable agent already solves nearly all of them, so the pass rate has little room to move.

We built the tasks from merged pull requests instead, each reverted at a single frozen base commit. They are harder, and the baseline resolve rate leaves room for a SKILL to matter. But a repository yields few of them, and each one costs a full agent run, so the splits stay small. The gains we measure --- $4.9$pp for GEPA and $0.1$pp for SkillOpt --- are not separable from the agent's own variance at that size.

The documents are more convincing than the numbers. A maintainer of one of the studied libraries\footnote{\texttt{koog}, an agentic framework for Kotlin.} found in them repository knowledge that is not available without reading its history, and with either SKILL the agent reached its goal faster and at lower cost. Whether this holds more broadly is the open question. Answering it needs cheaper rollouts, or an evaluation that does not rest on a pass rate, and a developer study with many reviewers where ours has one.

\section*{Responsible use statement}

A synthesized SKILL can read as authoritative while stating claims that are stale or over-general: \cref{sec:tasks} shows one that told the agent to work out which of two repository variants it was in, where there was only one repository seen at different points of its history, and \cref{sec:exp-devs} a maintainer finding general advice presented as repository knowledge. Any SKILL this pipeline produces should be reviewed before it is merged, like any other pull request. The paired score measures a change in one agent's behaviour on one mined dataset; it does not certify that the document is accurate.

\section*{Acknowledgments}

We thank Antonii Belyshev for the maintainer review of the SKILLs for the \texttt{koog} library, and Georgii Zorabov for the maintainer review of the SKILLs for the \texttt{tracy} library.

\bibliography{references}
\bibliographystyle{ACM-Reference-Format}

\newpage{}

\appendix

\section{Defects the LLM reverter introduces}\label{app:reversion-defects}

The reverter (\cref{sec:tasks}) rewrites an old pull request's revert against today's code, which is the only way to get yield out of a fast-moving repository. But it reverts the call site and forgets the code behind it, in two shapes the validation gate cannot see, since the gate only asks whether the reversion breaks a test\@. \emph{Unimported symbol}: the reversion drops an import that surviving code still needs, so the task repository does not compile, and the gate reads that compile break as proof the hidden test broke --- correctly, in the common case, but not here\@. \emph{Stranded fix}: the reversion deletes the only call to a private function while leaving the function itself, so the task repository ships with the answer written out and named for the job. A static check, run once per mined pool before any rollout, catches both: on an intermediate \texttt{koog} pool of $280$ tasks it rejected $56$ ($36$ unimported-symbol, $26$ stranded-fix, $6$ both), of which $36$ had already passed the dynamic validation gate.

\section{Cost and time of the two open koog issues}\label{app:issue-cost}

\Cref{tab:issues} reports the cost and time of the six runs behind \cref{sec:exp-devs}: each of the two issues, solved with no SKILL, with the GEPA document, and with the SkillOpt one. The patches themselves are printed in \cref{app:issue-patches}.

\begin{table}[h]
\centering
\caption{Two issues open in \texttt{koog} at the time of the experiment, each solved by the same agent under three configurations\@. \emph{API} is time spent in model calls, \emph{wall} is end to end.}
\label{tab:issues}
\small
\begin{tabular}{llrrr}
\toprule
issue & configuration & cost & API & wall \\
\midrule
\#1275 & no SKILL & \$6.06 & 8\,m\,21\,s & 14\,m\,59\,s \\
       & GEPA     & \$2.48 & 3\,m\,35\,s & 5\,m\,32\,s \\
       & SkillOpt & \$2.44 & 3\,m\,43\,s & 5\,m\,30\,s \\
\addlinespace
\#1354 & no SKILL & \$4.85 & 7\,m\,21\,s & 20\,m\,29\,s \\
       & GEPA     & \$3.10 & 4\,m\,52\,s & 7\,m\,38\,s \\
       & SkillOpt & \$4.20 & 5\,m\,33\,s & 8\,m\,24\,s \\
\bottomrule
\end{tabular}
\end{table}

\section{Extended limitations}\label{app:limitations-extended}

This appendix expands \cref{sec:limitations}: what each of the two mining checks drops, why a reverse-applied count is not yet a pool of graded tasks, and what a new repository has to supply.

\paragraph{How many tasks survive validation.}
A merged pull request is not yet a task. First, the implementation must reverse-apply at the single frozen base the SKILL will ship against. If the files it touched have since been renamed, moved, or rewritten, that historical diff no longer applies, and the change is dropped. Second, we revert the surviving change in an isolated environment and re-run the suite. A task is admitted only if tests that passed at the unmodified base now fail, and those failures isolate that change: the FAIL\_TO\_PASS set has to be the specification of that one reversion. The candidate is dropped if no test changes from passing to failing; if the tests that would grade the change were already failing at the unmodified base; or if the reversion also fails tests that do not belong to the change. There is no numeric cutoff on how many extra tests may fail --- once failures spill past the change, the task no longer grades that change. A repository can fail this second check without any rename: a change its own tests never covered, or a main branch that is already red, leaves nothing for validation to mark as FAIL\_TO\_PASS.

\paragraph{A reverse-applied count is not yet a pool of graded tasks.}
Reverse-application asks only whether the historical diff still applies at the frozen tree. It does not ask whether reverting that change isolates tests. A new repository can therefore list hundreds of pull requests, reverse-apply a three-digit set of them --- including changes that need the LLM reverter because the files have drifted, and changes that shipped no test hunk of their own --- and still fall short of $100$ graded tasks once validation runs. The candidates without a test hunk are not yet lost: they can still become tasks if existing tests fail on revert, and only the second check can tell.

We saw this on \ghrepo{http4k/http4k}~\citep{http4k}, which we tried as a fourth repository --- an HTTP library --- so that a finished run would show the method is not tied to kotest, ktor, and koog. From $700$ listed pull requests we reverse-applied $150$ changes at a frozen base: $35$ with \texttt{git apply}, $82$ with the LLM reverter, and $33$ with no test hunk of their own. Validation then ran on those $150$ candidates and left fewer than $100$ graded tasks.

\paragraph{When the empty SKILL already solves the training tasks.}
We encountered the pass-rate ceiling of \cref{sec:limitations} with SkillOpt on http4k, before that collection ran. There was no validated pool yet, so the train set was four pull requests collected by hand. For each, the score was $1$ if the agent passed every hidden test and $0$ otherwise, and the number SkillOpt used to keep or reject a rewrite was the mean of those four scores. The empty SKILL already scored $1.0$ on that mean, so every rewrite was rejected and the run ended on the empty file.

\paragraph{What this means for a new repository.}
The two checks require three properties of the repository. The files those historical diffs touch must still be present in the frozen tree, otherwise reverse-application drops the change. The test suite must pass at the unmodified base, otherwise there is no FAIL\_TO\_PASS set to isolate. And reverting each surviving change must make some of those passing tests fail, with the failures confined to that one reversion. A linked issue is not among the requirements: when none exists we synthesise a leak-checked statement from the pull request (\cref{sec:tasks}).

On the repositories that met the bound, a few hundred recent merged pull requests were enough: after both checks, \texttt{ktor} retained $131$ graded tasks from $452$ recent merged pull requests and \texttt{koog} $119$ from $660$. We collected the most recent merged pull requests of each and stopped once $100$ graded tasks remained (\cref{sec:exp-setup}), so those counts are what these repositories yielded, not a rate a new one should expect.

\section{Objectives we rejected}\label{app:rejected}

This appendix is the sweep behind the pairing choice in \cref{sec:objective}. It is a preliminary run on \ghrepo{JetBrains/tracy}~\citep{tracy}, not the paired kotest, ktor, or koog results. The seed is the generic checklist in \cref{app:skill-seed} --- read the README, run a small test first, keep the diff small --- not the empty file of the reported protocol. Several of the selected SKILLs are that checklist with Tracy notes added. Binary resolve and the skill-blind judge shipped no document.

Each family rewards a different factor of the document or of the agent's work. We keep pairing because our question is whether the agent does better with the file than without it. The other families remain usable if a project wants a different factor; they write a different kind of file.

\Cref{tab:rejected-scores} measures each score on its own $0$--$1$ scale. \Cref{tab:rejected-compare} then rescores the shipped files on one shared six-task pool under a $14$-turn cap. The last row of \cref{tab:rejected-scores} is pairing, the score we keep: on this pool it shipped the seed.

\begin{table}[t]
\centering
\caption{Each family on its own $0$--$1$ scale, on a preliminary Tracy sweep. \emph{Empty} is the score of the seed document. \emph{Best} is the candidate the optimizer kept. \emph{Change} is held-out best minus held-out empty. A run marked ``no file'' shipped no advice. The last row is pairing. These numbers are not the paired results in \cref{sec:experiments}.}
\label{tab:rejected-scores}
\small
\begin{tabular}{@{}lccccc@{}}
\toprule
 & \multicolumn{2}{c}{Validation} & \multicolumn{2}{c}{Held-out} & \\
\cmidrule(lr){2-3}\cmidrule(lr){4-5}
Score & Empty & Best & Empty & Best & Change \\
\midrule
\multicolumn{6}{@{}l}{\emph{120 pull requests, 24 held-out tasks}} \\
Skill-reading ensemble & 0.084 & 0.117 & 0.112 & 0.121 & $+0.010$ \\
Skill-blind judge (no file) & 0.035 & 0.063 & 0.046 & 0.075 & $+0.029$ \\
Gold match (no file) & 0.166 & 0.166 & 0.188 & 0.188 & $0$ \\
Partial credit (no file) & 0.108 & 0.108 & 0.108 & 0.108 & $0$ \\
\midrule
\multicolumn{6}{@{}l}{\emph{Binary resolve, 39 solvable tasks}} \\
Binary resolve (no file) & 0.143 & 0.143 & 0.222 & 0.222 & $0$ \\
\midrule
\multicolumn{6}{@{}l}{\emph{Clean-diff pool, 29 tasks}} \\
Gold match & 0.296 & 0.317 & 0.337 & 0.284 & $-0.053$ \\
Token price & 0.460 & 0.502 & 0.293 & 0.293 & $0$ \\
Work mix & 0.387 & 0.400 & 0.357 & 0.329 & $-0.027$ \\
File facts & 0.185 & 0.498 & 0.178 & 0.486 & $+0.308$ \\
File policy & 0.300 & 0.387 & 0.300 & 0.387 & $+0.087$ \\
Pairing (no file) & 0.500 & 0.496 & 0.500 & 0.500 & $0$ \\
\bottomrule
\end{tabular}
\end{table}

\begin{table}[t]
\centering
\caption{The files those runs shipped, each scored again on the same six Tracy tasks under a $14$-turn cap. \emph{Binary resolve} is $1$ only if every hidden test passed. \emph{Partial credit} is the mean test-outcome tier. \emph{File facts} and \emph{file policy} score the markdown, not the agent's patch. \emph{Mean} is the average of ten scores on this pool. \emph{Rank} is the mean rank across those ten (1 = best of eight files). Higher is better, except for rank. The file-facts and tool-call documents produced no patch on this pool.}
\label{tab:rejected-compare}
\scriptsize
\begin{tabular}{@{}lcccccc@{}}
\toprule
Shipped file & Binary & Partial & File & File & Mean & Rank \\
 & resolve & credit & facts & policy & & \\
\midrule
Work-mix (\cref{app:skill-efficiency}) & 0.333 & 0.467 & 0.503 & 0.274 & 0.376 & 2.30 \\
Token-price (\cref{app:skill-cost}) & 0.333 & 0.467 & 0.339 & 0.373 & 0.358 & 3.50 \\
Generic seed (\cref{app:skill-seed}) & 0.333 & 0.450 & 0.194 & 0.314 & 0.313 & 3.70 \\
Gold-match (\cref{app:skill-reference}) & 0.333 & 0.467 & 0.514 & 0.250 & 0.322 & 3.80 \\
File-facts (\cref{app:skill-gmu}) & 0.333 & 0.467 & 0.531 & 0.233 & 0.267 & 4.80 \\
Ensemble (\cref{app:skill-hermes}) & 0.167 & 0.308 & 0.502 & 0.233 & 0.288 & 5.70 \\
Tool-call (\cref{app:skill-deveffort}) & 0.333 & 0.467 & 0.503 & 0.186 & 0.265 & 5.70 \\
File-policy (\cref{app:skill-ideal}) & 0.167 & 0.300 & 0.327 & 0.404 & 0.230 & 6.50 \\
\bottomrule
\end{tabular}
\end{table}

\paragraph{Counting solved tasks.}
Binary resolve gives a point only if every hidden FAIL\_TO\_PASS test passed. About $92\%$ of those rollouts scored $0$, so the optimizer almost never saw a gradient and shipped the seed. Partial credit gives some points when the build compiled but the tests still failed. SkillOpt's hard gate accepts a rewrite only when the agent fully passed more tasks than before; on a tiny train set the seed can sit at $0$ or at $1.0$, and the gate then freezes. When a solved-task score did write a document, it wrote the training pull requests back into prose. \Cref{app:skill-test} tells the agent how to convert \texttt{tracing/test-utils} to \texttt{java-test-fixtures}, where to bump \texttt{okhttp}, and how \texttt{LangfuseMediaSpanProcessor} should upload media. That is useful for the instances the optimizer saw. It is not a reusable policy. We drop the family because the number ranks how easy the split is, and when it moves it overfits the harness.

\paragraph{Matching the gold patch.}
CodeBLEU, overlap of the assertions that changed, and exact match ask how close the agent's patch is to the change the maintainer merged. The document they selected is \cref{app:skill-reference}: a Tracy module list, \texttt{TracingManager.setSdk}, and \texttt{@KotlinFlowTrace}, written so the agent will touch the same files as the gold patch. A correct fix written differently is punished. On Kotlin the CodeBLEU we can compute is mostly $n$-gram overlap, not the AST and data-flow terms the CodeBLEU paper describes. Ten runs of the same SKILL on the same already-solved task still spanned CodeBLEU $0$ to $1$. On that protocol the held-out score fell (\cref{tab:rejected-scores}).

\paragraph{Pricing the run.}
Mixing ``did the tests pass'' with token or dollar cost makes the empty SKILL a local optimum, because an empty file adds no extra prompt tokens. When that score did not collapse to an empty file, it selected \cref{app:skill-cost}: which modules use \texttt{jvmTest} and which use \texttt{test}. Pricing tool calls writes the opposite document. \Cref{app:skill-deveffort} is a full Tracy map --- directory tree, \texttt{TracingManager} paths, per-module \texttt{./gradlew} lines --- so the agent will explore less. A work-priced mix selected the leaner guide in \cref{app:skill-efficiency}. Both horns are wrong for a SKILL: the cheap end ships no advice, and the expensive end ships a map the agent then spends its turns reading. Under a $14$-turn cap the long document emitted no patch.

\paragraph{Asking a judge.}
A judge that sees only the patch and the problem statement never reads the SKILL, so there is no document to ship. On one confirmation run the held-out score rose while the file was empty, because the judge was scoring the agent's writing. An ensemble that does read the SKILL, and asks whether the agent followed it, selected \cref{app:skill-hermes}: repository identity, the top-level tree, and a readable account of \texttt{TracingManager}. That is an onboarding note a human can follow. It never asks whether the agent did better with the SKILL than without it.

\paragraph{Scoring the file.}
These scores ask whether the markdown looks like a good SKILL. The \texttt{gmu} score counts verifiable repository facts. It selected \cref{app:skill-gmu}: the real module list, and the rule that \texttt{ai-dev-kit-tracing} must be tested with \texttt{jvmTest}. The \texttt{ideal} score measures completeness and operational tone. It selected \cref{app:skill-ideal}: compile and test playbooks, and the rule that a \texttt{@KotlinFlowTrace} attribute handler must be a Kotlin \texttt{object}. Both files look like something a maintainer might ship. The score is a property of the file, not of the agent. Grounding can rise on held-out tasks while the agent does not improve. On the shared pool the \texttt{gmu} document emitted no patch, and the \texttt{ideal} document was the worst of the set on every metric that looked at behaviour (\cref{tab:rejected-compare}).

\paragraph{Why pairing.}
Pairing asks a causal question: does this document change what the agent does, on this task, relative to the seed? Counting solved tasks answers how easy the repository is. Matching the gold patch answers whether the wording matches the maintainer. Pricing the run answers how cheap the attempt was. A judge answers how the trace reads. Scoring the file answers whether the markdown looks complete. Those last two can move while the agent does not, which is why \cref{app:skill-gmu,app:skill-ideal} rose on held-out tasks and still lost on behaviour. Against an empty seed, $0.5$ means the document changed nothing. On the same split used for \texttt{ideal}, pairing shipped the seed. That is the test we wanted.

The listings below are the documents those scores selected.

\subsection{Generic seed}\label{app:skill-seed}

This is the short, repository-agnostic checklist used as the Tracy sweep baseline in \cref{tab:rejected-compare}. The paired runs start from an empty file instead.

\begin{lstlisting}[style=skillmd]
---
name: repo-skill
description: Repository-specific guidance for working in this codebase. Always consult before making changes.
---

# Working in this repository

You are coding inside this repository. Use the following defaults until you discover better practices by inspecting the code:

## Discover before editing
- Read the project's README, CONTRIBUTING.md, and any AGENTS.md / CLAUDE.md before non-trivial edits.
- Use Glob and Grep to locate relevant code paths instead of guessing file names.
- When you change a public API, search for all callers first.

## Run tests early and iterate from failures
- Find the test runner from the build manifest (`pyproject.toml` / `requirements*.txt` for Python; `build.gradle.kts` / `pom.xml` for JVM) before guessing commands.
- Run a small, fast subset first (one test file or one module) before running the full suite.
- When a test fails, read the full traceback and the failing test source before changing code.

## Make minimal changes
- Match existing style, naming, and abstractions. Don't introduce new patterns unless asked.
- Prefer fixing the root cause over adding compensating logic elsewhere.
- Don't touch unrelated files.
\end{lstlisting}

\subsection{Solved-task skill}\label{app:skill-test}

Selected by partial-credit test resolve. Task-specific Tracy recipes: a \texttt{java-test-fixtures} conversion, version bumps, and \texttt{LangfuseMediaSpanProcessor}.

\begin{lstlisting}[style=skillmd]
# Current best skill (cand=f4d3cbca)
# mean=0.833  n=3  sum=2.50  cost=$4.24
# per-task: 146=1.0, 182=0.5, 205=1.0
# from iter 4 (GEPA accepted into pareto front)

---
name: repo-skill
description: Repository-specific guidance for working in this Kotlin/JVM tracing codebase (JetBrains Tracy / ai.jetbrains.tracy / org.jetbrains.ai.tracy). Always consult before making changes.
---

# Working in this repository

This is a multi-module Gradle Kotlin project (JetBrains "Tracy" tracing library) with modules under `tracing/` (e.g. `tracing/core`, `tracing/openai`, `tracing/anthropic`, `tracing/test-utils`, `tracing/ktor`). Build files are `build.gradle.kts`. Tests live under `src/jvmTest/kotlin` and main sources under `src/jvmMain/kotlin`. Package roots include `ai.jetbrains.tracy.*` and `org.jetbrains.ai.tracy.*`.

## Discover before editing
- Always start by reading `README.md`, `CONTRIBUTING.md`, `AGENTS.md`, `CLAUDE.md` (if present) and the root `build.gradle.kts` / `settings.gradle.kts` to learn module layout and conventions.
- Use Glob and Grep aggressively to locate relevant code; do not guess file names. Useful patterns:
  - `find . -path "*/jvmMain/*" -name "*.kt" | xargs grep -l <symbol>`
  - `find . -path "*/jvmTest/*" -name "*.kt" | xargs grep -l <symbol>`
  - Glob `**/*<ClassName>*` to find a class and its tests.
- When changing a public API, search for ALL callers across every module first, including example/client modules. Cross-module breakage (e.g. `tracing/anthropic` referencing something added in `tracing/test-utils`) is a common failure mode.

## Gradle / Kotlin specifics for this repo
- This is a Kotlin Multiplatform project; most JVM code lives in `jvmMain` / `jvmTest` source sets.
- When converting a module to Java test fixtures (e.g. `test-utils`):
  - Apply the `java-test-fixtures` plugin in the producing module.
  - In every consuming module's `build.gradle.kts`, replace `implementation(project(":tracing:test-utils"))` (or similar) with `implementation(testFixtures(project(":tracing:test-utils")))` in the appropriate source set / dependency block. With Kotlin Multiplatform, `testFixtures(...)` is NOT available inside `kotlin { sourceSets { ... } }` dependency blocks directly — it must be referenced from a `dependencies { ... }` block where Java's `testFixtures` helper is in scope. Verify by compiling; "Unresolved reference: testFixtures" means the dependency block is wrong context.
  - Search every `build.gradle.kts` in the repo for references to the converted module and update them all. Missing one consumer (e.g. `tracing/anthropic`) breaks the whole build.
  - Move shared test helpers (e.g. `createTestSpanData`) from individual test source sets into `test-utils` testFixtures source set and delete duplicates.
- Kotlin version bumps (e.g. to `2.1.21`) are typically in `gradle/libs.versions.toml` or root `build.gradle.kts`.
- Library version bumps (e.g. okhttp `4.12.0` -> `5.3.2`, adding `okhttp-coroutines` for `Call.executeAsync`) go in `libs.versions.toml`.

## Domain knowledge
- `ContentCapturePolicy` / `TracingManager`: capturing policies redact user/model input/output content in spans. OpenAI handlers (chat completions, responses, images) and Anthropic handlers must respect the policy. Tests extend `BaseOpenTelemetryTracingTest`, which exposes a `withCapturingPolicy` / `traceSensitiveContent` toggle.
- `protocol` package abstracts HTTP types (`Response`, `Request`, `ContentType`, `Url`) as interfaces so Ktor and OkHttp implementations can coexist; the migration removes Ktor from `core`.
- `LangfuseMediaSpanProcessor` uploads media via OkHttp (PUT to `uploadUrl`); `uploadMediaFileToLangfuse` should log errors on failure. Active job tracking must be guarded by `synchronized` blocks and must await jobs added between snapshot and await.
- `isStreamingRequestKey` should be set to `false` when `req` is null.
- `DataUrl` and its tests should not depend on Ktor after the migration.

## Run tests early and iterate from failures
- Discover the test runner from `build.gradle.kts` / `libs.versions.toml`. Standard commands:
  - Single module test: `./gradlew :tracing:<module>:jvmTest --tests "<FQCN>"`
  - Compile check first: `./gradlew :tracing:<module>:compileKotlinJvm` and `./gradlew :tracing:<module>:compileTestKotlinJvm`
  - Build script compilation: `./gradlew help` (catches `build.gradle.kts` errors like "Unresolved reference: testFixtures").
- Run the build-script compile / a fast subset before the full suite. Always re-run after editing any `build.gradle.kts`.
- Read the full traceback and the failing test source before editing production code.

## Make minimal, consistent changes
- Match existing style, naming, package layout, and abstractions.
- Fix root causes; don't add compensating logic elsewhere.
- Don't touch unrelated files, but DO update every consumer when a shared module's API changes (test-utils, protocol interfaces, version bumps). Incomplete propagation across modules is the most common cause of test failures here.
\end{lstlisting}

\subsection{Gold-match skill}\label{app:skill-reference}

Selected by CodeBLEU and gold-patch match. Modules and APIs aligned with the maintainer's patch.

\begin{lstlisting}[style=skillmd]
---
name: repo-skill
description: Repository-specific guidance for working in this codebase. Always consult before making changes.
---

# Working in this repository

## Discover before editing
- Read README.md, CONTRIBUTING.md, AGENTS.md, and any `.claude/` or `.teamcity/` files before non-trivial edits.
- Use Glob and Grep to locate relevant code paths instead of guessing file names.
- When you change a public API, search for all callers first.

## Repository structure

This is the **Tracy** AI tracing library for Kotlin/Java by JetBrains. The root project name is `tracy`.

Key submodules (declared in `settings.gradle.kts`):
- `tracing/core` — Kotlin Multiplatform (commonMain + jvmMain + jvmTest); core OpenTelemetry tracing
- `tracing/openai`, `tracing/anthropic`, `tracing/gemini`, `tracing/ktor` — JVM-only adapter modules
- `tracing/test-utils` — shared test helpers
- `eval` — JVM evaluation framework (not multiplatform)
- `examples` — runnable usage examples
- `plugin/gradle-tracy-plugin` and `plugin/tracy-compiler-plugin-*` — Gradle and compiler plugins
- `docs` — MkDocs documentation

Source layout conventions:
- Multiplatform modules: `src/commonMain/kotlin/`, `src/jvmMain/kotlin/`, `src/jvmTest/kotlin/`
- JVM-only modules: `src/main/kotlin/`, `src/test/kotlin/`
- Package root: `ai.dev.kit` (most modules); `ai.jetbrains.tracy` (openai module)

## Key APIs

### Tracing core (`ai.dev.kit.tracing.fluent`)
- `withSpan(name, attributes, block)` — inline function wrapping a block in an OpenTelemetry span; `withTrace()` is deprecated, use `withSpan()` instead
- `TracingManager` — singleton managing the OTel SDK; configure via `TracingManager.setSdk(...)` and `configureOpenTelemetrySdk(exporterConfig)`
- `@KotlinFlowTrace(name, spanType)` — runtime-retained annotation marking methods for automatic tracing
- `FluentSpanAttributes` — enum of MLflow/OTel span attribute keys
- `SpanType` — constants: `LLM`, `CHAIN`, `AGENT`, `TOOL`, `CHAT_MODEL`, `RETRIEVER`, `PARSER`, `EMBEDDING`
- `SpanMetadataCustomizer` / `DefaultSpanMetadataCustomizer` — expect/actual pattern for platform-specific span metadata
- `TracingSessionProvider.currentSessionId` / `withSessionIdBlocking` — session ID management
- Follow OpenTelemetry GenAI semantic conventions (`io.opentelemetry.semconv.incubating.GenAiIncubatingAttributes`); do not invent custom attribute names

### Eval framework (`ai.dev.kit.eval.utils`)
- `Generator<AIInputT, AIOutputT>` — interface with `suspend fun generate(input): AIOutputT`
- `Evaluator<GroundTruthT, AIOutputT, EvalResultT>` — interface with `evaluate()` and optional `aggregateResults()`
- `TestCase<AIInputT, GroundTruthT>` — data class holding `name`, `input`, `groundTruth`
- `EvalResult` — interface; `SingleScoreEvalResult`, `MultiScoreEvalResult` are concrete implementations
- `BaseEvaluationTest` — abstract JUnit 5 base class managing runs, tracing, and logging
- `ConsoleEvaluationTest` (in `NoLoggingEvaluationTest.kt`) — subclass with no external logging
- `LoggingClient` — interface for experiment/run/metric backends
- `LangfuseEvaluationTest` — combines `BaseEvaluationTest` with Langfuse via `LangfuseEvaluationClient`

### Security conventions (from AGENTS.md)
- Never commit API keys or credentials; use environment variables for sensitive configuration
- The system redacts AI inputs/outputs by default — preserve this behaviour, do not remove it
- Spans only emit when the SDK is installed and tracing is enabled; maintain minimal overhead when disabled

## Build and test

- **Build system:** Gradle with Kotlin DSL; use `./gradlew` from the repository root
- **Run tests:** `./gradlew test` (full suite) or target a single module, e.g. `./gradlew :tracing:core:jvmTest`
- **Find the test runner** from `build.gradle.kts` / `settings.gradle.kts` before guessing commands
- If a build fails with "Timeout waiting to lock journal cache", another Gradle daemon is running — wait or stop it before retrying
- Run a small, fast subset first (one module or one test file) before running the full suite
- When a test fails, read the full traceback and the failing test source before changing code

## Make minimal changes
- Match existing style, naming, and abstractions; do not introduce new patterns unless asked
- Prefer fixing the root cause over adding compensating logic elsewhere
- Do not touch unrelated files
- Justify all new dependencies; add tests for new functionality
\end{lstlisting}

\subsection{Token-price skill}\label{app:skill-cost}

Selected by token-priced cost, when that score did not collapse to an empty document. Compact Gradle test-task conventions.

\begin{lstlisting}[style=skillmd]
---
name: repo-skill
description: Repository-specific guidance for working in this codebase. Always consult before making changes.
---

# Working in this repository

## Discover before editing
- Read the project's README, CONTRIBUTING.md, and any AGENTS.md / CLAUDE.md before non-trivial edits.
- Use Glob and Grep to locate relevant code paths instead of guessing file names.
- When you change a public API, search for all callers first.

## Run tests early and iterate from failures
- Find the test runner from the build manifest (`pyproject.toml` / `requirements*.txt` for Python; `build.gradle.kts` / `pom.xml` for JVM) before guessing commands.
- Run a small, fast subset first (one test file or one module) before running the full suite.
- When a test fails, read the full traceback and the failing test source before changing code.

## Make minimal changes
- Match existing style, naming, and abstractions. Don't introduce new patterns unless asked.
- Prefer fixing the root cause over adding compensating logic elsewhere.
- Don't touch unrelated files.

## Gradle test task conventions
The correct test task depends on which Gradle plugin a module uses:

| Plugin | Test task |
|---|---|
| `kotlin("jvm")` / `kotlin.jvm` | `test` |
| `kotlin("multiplatform")` / `kotlin.multiplatform` | `jvmTest` |

Using the wrong task causes a "task not found" build failure. Check the module's `build.gradle.kts` for the applied plugin before running tests.

```bash
# JVM-only module
./gradlew :some-module:test

# Multiplatform module
./gradlew :some-module:jvmTest
```

## Source layout conventions
- **Multiplatform modules:** `src/commonMain/kotlin/`, `src/jvmMain/kotlin/`, `src/jsMain/kotlin/`, `src/jvmTest/kotlin/`
- **JVM-only modules:** `src/main/kotlin/`, `src/test/kotlin/`
- **Test fixtures:** `src/testFixtures/kotlin/`

## Kotlin visibility and compilation errors
Common compilation errors to watch for:
- A `private` member accessed from another class in the same package — fix by widening to `internal`.
- An `internal` property with a `private` setter accessed from outside the object — the getter visibility must also permit access.
- An unresolved reference that looks like a property name — check whether the property actually exists on the referenced object; it may have been renamed or never existed.

Always verify fixes compile before running tests:
```bash
./gradlew compileKotlin          # JVM-only module
./gradlew compileKotlinJvm       # multiplatform module
```

## JUnit dependency completeness
If tests use `@ParameterizedTest` or `@MethodSource`, the `junit-jupiter-params` artifact must be explicitly on the test classpath — it is not pulled in transitively by `kotlin-test-junit5`. Add it explicitly:
```kotlin
testImplementation("org.junit.jupiter:junit-jupiter-params:<version>")
```
\end{lstlisting}

\subsection{Tool-call-effort skill}\label{app:skill-deveffort}

Selected by tool-call effort (a non-empty fallback). A full Tracy map with per-module \texttt{./gradlew} lines.

\begin{lstlisting}[style=skillmd]
---
name: repo-skill
description: Repository-specific notes learned while working in this codebase.
---

## Repository Identity

**Tracy** — a JetBrains AI Tracing Library for Kotlin/Java. Integrates with OpenTelemetry and observability platforms (Langfuse, Weights & Biases).

## Top-Level Layout

```
/workspace/repo/
├── buildSrc/          # Shared Gradle convention plugins
├── docs/              # MkDocs documentation
├── eval/              # Evaluation module (JVM only, kotlin("jvm"))
├── examples/          # Runnable examples
├── plugin/            # Kotlin compiler plugins (versions 1.9.0, 1.9.20, 2.1.0, 2.2.20) + gradle-tracy-plugin
├── publishing/        # Artifact publishing plugin (composite build)
├── tracing/           # Main library modules
├── build.gradle.kts
├── gradle.properties
├── gradle/libs.versions.toml
└── settings.gradle.kts
```

## Gradle Modules

Declared in `settings.gradle.kts`:
```
include("eval")
include("examples")
include("tracing:core")
include("tracing:anthropic")
include("tracing:gemini")
include("tracing:ktor")
include("tracing:openai")
include("tracing:test-utils")
includeBuild("plugin")
includeBuild("publishing")
```

All `tracing:*` modules use `kotlin("multiplatform")` targeting JVM 17. `eval` uses `kotlin("jvm")`.

**Always run `./gradlew projects` to confirm available subproject names before constructing task paths.**

## Source Layout Convention (Kotlin Multiplatform)

Each `tracing/*` module follows:
```
src/
  commonMain/kotlin/   # Shared expect declarations
  jvmMain/kotlin/      # JVM actual implementations
  jsMain/kotlin/       # JS actual implementations
  jvmTest/kotlin/      # JVM tests
  jvmTestFixtures/kotlin/  # Java test fixtures (test-utils module)
```

## Key Module: `tracing/core`

- `src/jvmMain/kotlin/ai/dev/kit/`
  - `OpenTelemetryOkHttpInterceptor.kt` — `patchOpenAICompatibleClient<Client, ClientImpl, ClientOptions, ClientOkHttpClient>()` injects a custom OkHttp interceptor via reflection
  - `adapters/LLMTracingAdapter.kt` — abstract base for LLM tracing adapters; sets `gen_ai.api_base`, `gen_ai.system` span attributes
  - `adapters/handlers/EndpointApiHandler.kt` — interface with `handleRequestAttributes(span, request)` and `handleResponseAttributes(span, response)`
  - `tracing/TracingManager.kt` — singleton `object`; controls `isTracingEnabled`, holds `ContentCapturePolicy`
- `src/commonMain/kotlin/ai/dev/kit/tracing/fluent/`
  - `KotlinFlowTrace.kt` — `@KotlinFlowTrace` annotation
  - `FluentSpanAttributes.kt` — enum of span attribute keys
  - `TracingSessionProvider.kt`, `TracingMetadataConfigurator.kt`, `SpanType.kt`
  - `handlers/SpanAttributeHandler.kt`, `handlers/BaseSpanAttributeHandler.kt`

## Key Module: `tracing/openai`

- Package root: `ai/dev/kit/` (or `ai/jetbrains/tracy/tracing/` depending on repo variant)
- `adapters/OpenAILLMTracingAdapter.kt` — routes to handler based on request body keys (`"messages"` → ChatCompletions, `"input"` → Responses, etc.)
- `adapters/openai/` or `adapters/handlers/` — `ChatCompletionsHandler`, `ResponsesApiHandler`, image handlers
- `clients/OpenAIClient.kt` — `instrument(client: OpenAIClient): OpenAIClient` is the recommended entry point for adding tracing

## Key Module: `tracing/anthropic`

- `clients/AnthropicAIClient.kt` — `instrument(client: AnthropicClient): AnthropicClient`
- Pattern for `instrument()`: calls `patchOpenAICompatibleClient(...)` from `tracing-core` with a custom interceptor

## Key Module: `tracing/test-utils`

- Sources live under `src/jvmTestFixtures/kotlin/ai/jetbrains/tracy/test/utils/`
- `BaseOpenTelemetryTracingTest` — sets up `InMemorySpanExporter`, calls `TracingManager.setSdk(...)`, enables sensitive content capture in `@BeforeTest`
- `BaseAITracingTest` — extends `BaseOpenTelemetryTracingTest`; provides `validateBasicTracing(url, model)`, `provideContentCapturePolicies()`
- `MediaSource` — sealed class for file/link media sources

## Build and Test Commands

```bash
# Build entire project
./gradlew build

# Run all non-local tests
./gradlew test

# Run tests for a specific module
./gradlew :tracing:openai:jvmTest
./gradlew :tracing:anthropic:jvmTest
./gradlew :tracing:core:jvmTest
./gradlew :tracing:test-utils:check

# Compile only (useful to check for errors without running tests)
./gradlew :tracing:openai:compileTestKotlinJvm
./gradlew :tracing:test-utils:compileKotlinJvm

# List all available subprojects
./gradlew projects
```

## Test Configuration

- All `Test` tasks use JUnit Platform
- Tests tagged `SkipForNonLocal` are excluded unless `-DaiDevKitLocalTests=true` (or `-DtracyLocalTests=true`) is passed
- Integration tests requiring live endpoints (e.g. `LITELLM_URL`, `OPENAI_API_KEY`) are tagged accordingly and excluded from CI by default

## Content Capture Policy

- `ContentCapturePolicy` data class in package `ai.dev.kit.tracing.policy` with fields `captureInputs: Boolean`, `captureOutputs: Boolean`
- `TracingManager.withCapturingPolicy(policy)` — sets active policy
- `TracingManager.traceSensitiveContent()` — enables full capture
- When `captureInputs == false`, prompt/tool input attributes are set to `"REDACTED"`; when `captureOutputs == false`, completion/output attributes are set to `"REDACTED"`

## OpenTelemetry Attribute Conventions

- Uses `GenAiIncubatingAttributes` constants: `GEN_AI_REQUEST_MODEL`, `GEN_AI_REQUEST_TEMPERATURE`, `GEN_AI_USAGE_INPUT_TOKENS`, `GEN_AI_USAGE_OUTPUT_TOKENS`
- Custom string attributes: `"gen_ai.api_base"`, `"gen_ai.system"`, `"gen_ai.prompt.$index.role"`, `"gen_ai.prompt.$index.content"`, `"gen_ai.completion.$index.content"`, `"gen_ai.tool.$index.type/name/description/parameters"`
- `FluentSpanAttributes` keys: `SPAN_INPUTS("input")`, `SPAN_OUTPUTS("output")`, `SOURCE_RUN("session.id")`, `SPAN_TYPE("spanType")`, etc.

## Version Catalog (`gradle/libs.versions.toml`) — Key Versions

```toml
kotlin = "2.1.0"          # (may be updated to 2.1.21)
kotlinx-coroutines = "1.9.0"
kotlinx-serialization = "1.5.1"
okhttp = "4.12.0"
opentelemetry = "1.51.0"
opentelemetry-semconv-incubating = "1.34.0-alpha"
junit = "5.10.0"
```

## Convention Plugins (buildSrc)

- `id("ai.jetbrains.tracy.published-artifact")` — marks module for publishing
- `id("org.jetbrains.ai.tracy")` — Tracy compiler plugin integration
- `id("ai.kotlin.dokka")` — Dokka documentation generation
- `id("ai.dev.kit.trace")` — enables `@KotlinFlowTrace` compiler plugin for a module

## Known Pitfalls

- **Gradle lock contention**: if a build fails with "Timeout waiting to lock journal cache", wait for any other Gradle process to finish before retrying.
- **Subproject naming**: always verify subproject names with `./gradlew projects` — names like `ai-dev-kit-tracing-langfuse` vs `ai-dev-kit-tracking-langfuse` have caused errors.
- **`instrument()` functions**: each LLM provider module must expose its own `instrument(client)` function in `src/jvmMain`; tests import from the provider-specific package (e.g. `ai.dev.kit.clients`).
- **`@KotlinFlowTrace` handler**: the `attributeHandler` field must reference a Kotlin `object` singleton (accessed via `.objectInstance`).
- **Coroutine parameters**: `BaseSpanAttributeHandler.processInput` filters out `Continuation` parameters when building JSON from method arguments.
- **`-java-parameters` compiler flag**: required for parameter names to be visible at runtime for tracing.
\end{lstlisting}

\subsection{Work-mix skill}\label{app:skill-efficiency}

Selected by work-priced efficiency. A leaner module and build guide.

\begin{lstlisting}[style=skillmd]
---
name: repo-skill
description: Repository-specific guidance for working in this codebase. Always consult before making changes.
---

# Working in this repository

## Repository overview

This is the **AI Development Kit (Tracy)**, a JetBrains open-source Kotlin library for tracing, monitoring, and evaluating AI-powered features. It is built on OpenTelemetry and integrates with observability backends (Langfuse, W&B Weave, Jaeger, etc.).

Build system: **Gradle with Kotlin DSL** throughout. The root project is named `tracy`.

## Module layout

Submodules (from `settings.gradle.kts`):
- `tracing:core` — core tracing, Kotlin Multiplatform (commonMain + jvmMain + jsMain)
- `tracing:anthropic`, `tracing:gemini`, `tracing:openai`, `tracing:ktor` — AI client adapters
- `tracing:test-utils` — shared test helpers
- `eval` — evaluation framework (JVM only)
- `examples` — runnable examples
- `plugin/gradle-tracy-plugin`, `plugin/tracy-compiler-plugin-*` — compiler plugins

A `publishing/` directory is included as a composite build.

## Source set conventions

**tracing:core** uses Kotlin Multiplatform source sets:
- `commonMain` — shared API (`FluentSpanAttributes`, `KotlinFlowTrace`, `TracingSessionProvider`, `SpanMetadataCustomizer`, etc.)
- `jvmMain` — JVM `actual` implementations; `PlatformMethod` is `actual typealias PlatformMethod = java.lang.reflect.Method`
- `jsMain` — JS stubs; most functions call `TODO("Implementation depends on OpenTelemetry, which is JVM-only")`
- `jvmTest` — JVM tests under `src/jvmTest/kotlin/ai/dev/kit/tracing/fluent/`

**eval** module: `src/main/kotlin/ai/dev/kit/eval/utils/` and `src/main/kotlin/ai/dev/kit/eval/providers/`

**Package roots:**
- Core tracing: `ai.dev.kit.tracing.*`
- Eval framework: `ai.dev.kit.eval.utils.*`, `ai.dev.kit.eval.providers.*`
- AI client adapters: `ai.dev.kit.adapters.*`, `ai.dev.kit.clients.*`
- Examples: `ai.dev.kit.examples.*`

## Key APIs

### tracing:core — span creation

`withSpan` (in `processor/Utils.kt`, jvmMain):
```kotlin
inline fun <T> withSpan(
    name: String,
    attributes: Map<String, Any?> = emptyMap(),
    block: (Span) -> T
): T
```
Uses `TracingManager.tracer` to build, start, and close spans. Currently calls `.toString()` on all attribute values (`// TODO: deal with types`).

`@KotlinFlowTrace` — runtime-retention annotation that drives automatic span creation via the compiler plugin or interceptor.

`DefaultSpanMetadataCustomizer` — `expect object` in commonMain; JVM `actual` implementation in jvmMain.

### eval — evaluation framework

Core interfaces (package `ai.dev.kit.eval.utils`):
- `AIInput`, `AIOutput`, `GroundTruth` — marker interfaces
- `Generator<AIInputT, AIOutputT>` — wraps the AI feature under test; `suspend fun generate(input)`
- `Evaluator<GroundTruthT, AIOutputT, EvalResultT>` — `fun evaluate(...)` + optional `fun aggregateResults(...)`
- `TestCase<AIInputT, GroundTruthT>` — `name`, `input`, `groundTruth`
- `EvalResult`, `SingleScoreEvalResult`, `MultiScoreEvalResult` — result types
- `AggregateScore(scoreName: String, score: Double)` — data class
- `LoggingClient` — interface for experiment/run/metric logging; implemented by `LangfuseEvaluationClient`
- `BaseEvaluationTest` — abstract JUnit 5 base class for multi-run experiments
- `ConsoleEvaluationTest` (in `NoLoggingEvaluationTest.kt`) — `BaseEvaluationTest` subclass with no external logging

`getUserIDFromEnv()` reads the `USER_ID` environment variable.

## Discover before editing

- Read `README.md`, `CONTRIBUTING.md`, and any `AGENTS.md` / `CLAUDE.md` before non-trivial edits.
- Use Glob and Grep to locate relevant code paths instead of guessing file names.
- When you change a public API, search for all callers first — especially across `commonMain`/`jvmMain`/`jsMain` source sets.
- When a symbol is missing, check both `commonMain` and the platform-specific source set (`jvmMain`/`jsMain`) before adding it.

## Build and test commands

```bash
# Compile JVM sources for tracing:core
./gradlew :tracing:core:compileKotlinJvm

# Compile JVM test sources for tracing:core
./gradlew :tracing:core:compileTestKotlinJvm

# Run JVM tests for tracing:core
./gradlew :tracing:core:jvmTest

# Run tests for the eval module
./gradlew :eval:test

# Run a specific module's tests
./gradlew :<module>:jvmTest   # multiplatform modules
./gradlew :<module>:test      # JVM-only modules
```

If Gradle reports `Timeout waiting to lock journal cache`, another Gradle process is running — wait for it to finish before retrying.

## Run tests early and iterate from failures

- Run a small, fast subset first (one module) before running the full suite.
- When a test fails, read the full error output and the failing test source before changing code.
- For `compileTestKotlin` failures, read the test file to understand what symbols it imports before adding or modifying production code.

## Make minimal changes

- Match existing style, naming, and abstractions. Don't introduce new patterns unless asked.
- Prefer fixing the root cause over adding compensating logic elsewhere.
- Don't touch unrelated files or source sets.
- When adding a constant or function used by tests, place it in the same file/object where related constants already live.
\end{lstlisting}

\subsection{Skill-reading ensemble skill}\label{app:skill-hermes}

Selected by the judge ensemble that reads the SKILL. Onboarding note: repository identity, a directory tree, and \texttt{TracingManager}.

\begin{lstlisting}[style=skillmd]
---
name: repo-skill
description: Repository-specific notes learned while working in this codebase.
---

## Repository Identity

This is **Tracy**, a JetBrains AI tracing library for Kotlin and Java. The root project name is `tracy`.

## Top-Level Directory Structure

```
/workspace/repo/
├── buildSrc/          # Shared Gradle convention plugins
├── docs/              # MkDocs-based documentation site
├── eval/              # Evaluation module (LLM/AI evaluation utilities)
├── examples/          # Runnable example code
├── gradle/            # Version catalog (libs.versions.toml), wrapper, init scripts
├── plugin/            # Kotlin compiler plugins (multiple Kotlin versions) + Gradle plugin
├── publishing/        # Included build providing the publishing plugin
├── tracing/           # Core tracing library submodules
├── build.gradle.kts
├── gradle.properties
└── settings.gradle.kts
```

## Module Layout

### `tracing/` submodules
- `tracing/core` — core OpenTelemetry-based tracing abstractions
- `tracing/openai` — OpenAI adapter
- `tracing/anthropic` — Anthropic adapter
- `tracing/gemini` — Gemini adapter
- `tracing/ktor` — Ktor HTTP client adapter
- `tracing/test-utils` — shared test base classes (internal use only; structured as Java test fixtures)

### `plugin/` submodules
Compiler plugins for multiple Kotlin versions plus `gradle-tracy-plugin`.

## Package Naming

```
org.jetbrains.ai.tracy.<subproject-specific-suffix>
```
Provider-specific code uses: `ai.jetbrains.tracy.<provider>.*`  
Core/shared code may use: `ai.dev.kit.*` (legacy prefix, still present in some modules)

## Source Set Conventions (Kotlin Multiplatform)

All `tracing/` modules use Kotlin Multiplatform:
- `src/commonMain/kotlin/` — shared cross-platform code; `expect` declarations go here
- `src/jvmMain/kotlin/` — JVM `actual` implementations and JVM-specific code
- `src/jsMain/kotlin/` — JS `actual` implementations (often stubs throwing `NotImplementedError`)
- `src/jvmTest/kotlin/` — JVM tests
- `src/jvmTestFixtures/kotlin/` — Java test fixtures (used by `test-utils`)

When adding a new cross-platform abstraction:
1. Declare `expect class`/`expect object`/`expect fun` in `commonMain`.
2. Provide `actual` in both `jvmMain` (full implementation) and `jsMain` (stub or real).

JVM-only modules (e.g., `eval/`) use `src/main/kotlin/` and `src/test/kotlin/`.

## Build System

- **Build tool:** Gradle with Kotlin DSL
- **Version catalog:** `gradle/libs.versions.toml` — all dependency versions go here
- **JVM toolchain:** Java 17 (`jvmToolchain(17)`, `compilerOptions.jvmTarget = JVM_17`)
- `publishing/` is an included build via `pluginManagement { includeBuild("publishing") }` in `settings.gradle.kts`
- `buildSrc/` contains precompiled script convention plugins (e.g., `ai.kotlin.dokka.gradle.kts`)

### Running tests

| Module type | Task |
|---|---|
| Kotlin Multiplatform module | `./gradlew :tracing:core:jvmTest` |
| JVM-only module | `./gradlew :eval:test` |
| Check (compile + test) | `./gradlew :tracing:core:check` |

Only one Gradle daemon should run at a time; concurrent Gradle processes contend on `~/.gradle/caches/journal-1`.

### Test conventions
- JUnit 5 with `useJUnitPlatform()` throughout.
- Tests tagged `@Tag("SkipForNonLocal")` require live LLM endpoints and are excluded in CI.
- `BaseOpenTelemetryTracingTest` — abstract base class that sets up an `InMemorySpanExporter` and `TracingManager`; calls `TracingManager.traceSensitiveContent()` in `@BeforeTest` so span-content assertions remain valid.
- `BaseAITracingTest` — extends `BaseOpenTelemetryTracingTest` with AI-specific helpers.
- Consumer modules reference `test-utils` via `testFixtures()` notation:
  ```kotlin
  testImplementation(testFixtures(project(":tracing:test-utils")))
  ```

## Key Architectural Patterns

### OpenTelemetry tracing
- `TracingManager` (object in `tracing/core`) is the central singleton. Use `TracingManager.tracer` to obtain a `Tracer` — never call `GlobalOpenTelemetry.getTracer()` directly.
- `TracingManager.withCapturingPolicy(policy: ContentCapturePolicy)` controls content redaction.
- `TracingManager.traceSensitiveContent()` opts in to capturing both inputs and outputs (used in tests).
- `@KotlinFlowTrace(name, spanType, attributeHandler)` annotation instruments functions as OT spans (applied at compile time by the compiler plugin).

### LLM tracing adapters
- `LLMTracingAdapter` (in `tracing/core/src/jvmMain/`) is the abstract base for all provider adapters.
- Subclasses implement `getRequestBodyAttributes(span, url, body)` and `getResponseBodyAttributes(span, body)`.
- Span attribute constants come from `io.opentelemetry.semconv.incubating.GenAiIncubatingAttributes.*`.
- Each provider has an `instrument(client)` top-level function as the recommended entry point.

### Reflection-based interceptor injection
- `patchOpenAICompatibleClient` in `tracing/core/src/jvmMain/` uses reflection to inject an OkHttp interceptor into the SDK's internal client.
- **When upgrading OpenAI or Anthropic SDK versions**, verify the internal field names accessed by reflection (`okHttpClient`, `originalHttpClient`, `httpClient`) still match the new SDK class hierarchy.
- Both `OpenAIClient.kt` and `AnthropicAIClient.kt` follow the same pattern and must be kept in sync.

### expect/actual renaming
When renaming an `expect`/`actual` function or constant, update **all** of:
1. The `expect` declaration in `commonMain`
2. All `actual` implementations in `jvmMain` and `jsMain`
3. All call sites and imports in test files (`jvmTest`)
4. Any documentation referencing the old name

Use `grep -r 'OldName' .` to find all references before and after a rename.

## Dependency Management

All library versions are centralized in `gradle/libs.versions.toml`. Never hardcode versions in `build.gradle.kts` files — always add a version entry to the catalog and reference it via `libs.*` aliases.

Key aliases include: `libs.opentelemetry`, `libs.opentelemetry.semconv.incubating`, `libs.openai`, `libs.anthropic`, `libs.gemini`, `libs.okhttp`, `libs.kotlinx.serialization.core`, `libs.kotlinx.coroutines`.

## Module Registration

Every subproject must be registered in `settings.gradle.kts` with `include("path:to:module")` before it can be referenced as a Gradle project dependency. Run `./gradlew projects` to list all currently registered subprojects.

If an included build path changes (e.g., `includeBuild("publishing")`), the path in `settings.gradle.kts` must be updated to match, or all Gradle invocations will fail.

## Publishing

Published modules apply `id("ai.jetbrains.tracy.published-artifact")`. The `test-utils` module is **not** published externally. The publishing plugin is provided by the `publishing/` included build.

## Documentation

- User docs: MkDocs with Material theme, configured in `docs/mkdocs.yml`.
- API docs: Dokka (`ai.kotlin.dokka` convention plugin applied per module).
- Code snippets in docs are verified with Kotlin Knit (`docs/knit.properties`); generated sources go to `docs/src/`.
- To serve docs locally: `cd docs && uv sync --frozen --all-extras && uv run mkdocs serve`.
\end{lstlisting}

\subsection{File-facts skill (\texttt{gmu})}\label{app:skill-gmu}

Selected by grounding $\times$ density. A verifiable module list and the \texttt{jvmTest} versus \texttt{test} rule.

\begin{lstlisting}[style=skillmd]
---
name: repo-skill
description: Repository-specific guidance for working in this codebase. Always consult before making changes.
---

# Working in this repository

## Discover before editing
- Read the project's README, CONTRIBUTING.md, and any AGENTS.md / CLAUDE.md before non-trivial edits.
- Use Glob and Grep to locate relevant code paths instead of guessing file names.
- When you change a public API, search for all callers first.

## Run tests early and iterate from failures
- Find the test runner from the build manifest (`pyproject.toml` / `requirements*.txt` for Python; `build.gradle.kts` / `pom.xml` for JVM) before guessing commands.
- Run a small, fast subset first (one test file or one module) before running the full suite.
- When a test fails, read the full traceback and the failing test source before changing code.

## Make minimal changes
- Match existing style, naming, and abstractions. Don't introduce new patterns unless asked.
- Prefer fixing the root cause over adding compensating logic elsewhere.
- Don't touch unrelated files.

## Project structure

This repository is a multi-module Gradle project (Kotlin DSL). Modules observed across instances include:

```
ai-dev-kit-eval
ai-dev-kit-example
ai-dev-kit-plugin
ai-dev-kit-test-base
ai-dev-kit-tracing          ← Kotlin Multiplatform (JVM + JS)
ai-dev-kit-tracking-providers:ai-dev-kit-tracking-langfuse
ai-dev-kit-tracking-providers:ai-dev-kit-tracking-mlflow
ai-dev-kit-tracking-providers:ai-dev-kit-tracking-wandb
```

The tracing module may appear as `ai-dev-kit-tracing/` at the root or nested under `tracing/tracing-core/` — verify with `ls` or `find` before assuming a path.

## Critical: test task names differ by module type

**`ai-dev-kit-tracing` is a Kotlin Multiplatform module** — its JVM test task is `jvmTest`, not `test`:
```bash
./gradlew :ai-dev-kit-tracing:jvmTest
```

**All other modules** use the standard JVM plugin and the standard `test` task:
```bash
./gradlew :ai-dev-kit-eval:test
./gradlew :ai-dev-kit-test-base:test
./gradlew :ai-dev-kit-tracking-providers:ai-dev-kit-tracking-langfuse:test
./gradlew :ai-dev-kit-tracking-providers:ai-dev-kit-tracking-mlflow:test
./gradlew :ai-dev-kit-tracking-providers:ai-dev-kit-tracking-wandb:test
```

Never use `jvmTest` on `ai-dev-kit-test-base` or the tracking-provider modules — it will fail with "task not found".

## Test tag conventions

- `@Tag("SkipForNonLocal")` marks tests that require live external services (Langfuse, MLflow, W&B, LiteLLM). These are excluded in CI via the `aiDevKitLocalTests` system property set in the root `build.gradle.kts`.
- To run local-only tests: `./gradlew test -DaiDevKitLocalTests=true`
- All tests in the `ai-dev-kit-tracking-providers` submodules are tagged `SkipForNonLocal` and will not run in a standard offline environment.

## Module: `ai-dev-kit-tracing`

Kotlin Multiplatform module (JVM + JS). Source layout:

```
src/commonMain/kotlin/ai/dev/kit/tracing/fluent/   # expect declarations
src/jvmMain/kotlin/ai/dev/kit/tracing/fluent/      # actual JVM implementations
src/jsMain/kotlin/ai/dev/kit/tracing/fluent/       # JS stubs (most throw NotImplementedError)
```

### Key APIs

**`@KotlinFlowTrace`** — annotation for tracing functions; parameters: `name` (span name), `spanType`, `attributeHandler: KClass<out SpanAttributeHandler>`. Does not work with `suspend` functions in some configurations; check the README for current status.

**`TracingFlowProcessor`** (JVM object) — must call `setup()` / `setupTracing(di)` before tracing is active. Holds a `CoroutineScope(Dispatchers.IO + SupervisorJob())` and a Kodein `DI` container.

**`TracePublisher`** interface — `suspend fun publishTrace(trace: List<SpanData>)`

**`TracingSessionProvider`** (expect object) — `currentProjectId` and `currentSessionId`; use `withProjectId(id) { ... }` to scope tracing. JVM implementation uses OpenTelemetry `ContextKey`.

**`withTrace` / `withTraceSuspended`** (jvmMain) — retrieve `TracingMetadataConfigurator` from DI, require `@KotlinFlowTrace`, record exceptions with `span.recordException` and `StatusCode.ERROR`. Prefer `withSpan()` over the deprecated `withTrace()`.

**`RootSpanExporter`** — groups spans by `traceId`; publishes the complete trace when the root span (no valid `parentSpanId`) finishes.

**`FluentSpanAttributes`** enum (commonMain) — attribute keys including: `SPAN_INPUTS`, `SPAN_OUTPUTS`, `SPAN_FUNCTION_NAME`, `SPAN_SOURCE_NAME`, `SOURCE_RUN`, `TRACE_CREATION_INFO`, `SPAN_TYPE`, `TRACE_TAGS`. The `actual fun SpanData.getAttribute(...)` extension is in jvmMain. If you add a new enum entry, add it to `commonMain` and ensure jvmMain/jsMain `actual` implementations handle it.

**`BaseSpanAttributeHandler`** / `SpanAttributeHandler` — interface with `processInput(method, args)` and `processOutput(result)`. JVM actual serializes parameters to JSON (excluding `Continuation` parameters).

### Multiplatform expect/actual pattern

- `commonMain` holds `expect` declarations (interfaces, classes, functions, objects).
- `jvmMain` holds `actual` implementations backed by OpenTelemetry SDK and `java.lang.reflect`.
- `jsMain` holds stub `actual` implementations; most throw `NotImplementedError()` or `TODO()`.
- All source sets require the `-Xexpected-actual-classes` compiler argument.
- When adding a new `expect` declaration, provide `actual` implementations in both `jvmMain` and `jsMain`.

## Module: `ai-dev-kit-tracking-providers`

Three JVM-only provider submodules (Langfuse, MLflow, W&B). Each follows the same pattern:

- `*DiContainer` — Kodein DI object binding `TracePublisher` and `TracingMetadataConfigurator` as singletons.
- `Setup*Tracing.kt` — entry-point function (e.g. `setupLangfuseTracing(...)`) that calls `TracingFlowProcessor.setupTracing(di)`.
- `*TracePublisher` — reads `FluentSpanAttributes` from spans and POSTs to the provider's API.
- `*TracingMetadataConfigurator` — implements `TracingMetadataConfigurator`.

## Module: `ai-dev-kit-test-base`

JVM-only module providing abstract test fixture base classes consumed by provider test modules:

```
src/testFixtures/kotlin/ai/dev/kit/fluent/
├── TestFluentTracingBase.kt
├── TestSuspendFluentTracingBase.kt
└── TestAutologTracingBase.kt
```

## Module: `ai-dev-kit-eval`

Evaluation framework. Key types: `BaseEvaluationTest`, `EvaluationClient`, `Evaluator`, `Generator`, `TestCase`, `EvalResult`, `AggregateScore`, `EvalUtils`.

`BaseEvaluationTest` is annotated `@TestInstance(TestInstance.Lifecycle.PER_CLASS)` and integrates with OpenTelemetry and coroutines. Subclasses provide type parameters for input, output, ground truth, and eval result.

## Common compilation pitfalls

- **Private constants referenced across files**: if a constant is `private` in its declaring object, other files cannot access it. Change to `internal` or expose via a method.
- **`private set` on mutable state**: a field declared `internal var` with `private set` cannot be assigned from outside the declaring object, even from the same package. Remove `private set` or add an internal update function.
- **Missing enum entries**: if JVM code references a `FluentSpanAttributes` entry (e.g. `TRACE_TAGS`) that is absent from the `commonMain` enum, compilation fails with "Unresolved reference". Always add new entries to `commonMain`.
- **Unresolved references to mutable state**: verify that fields like `currentRunId` or `currentExperimentId` are actually declared in the object before referencing them.
- **Missing test dependencies**: `junit-jupiter-params` is not always transitively included; add it explicitly if using `@ParameterizedTest`.
- **`-java-parameters` compiler flag**: required to make method parameter names visible at runtime for tracing annotation processing.
- **JS stubs**: jsMain `actual` implementations intentionally throw; do not add real logic there unless explicitly targeting JS.
\end{lstlisting}

\subsection{File-policy skill (\texttt{ideal})}\label{app:skill-ideal}

Selected by the gated policy mix. Compile and test playbooks and the \texttt{@KotlinFlowTrace} \texttt{object} rule.

\begin{lstlisting}[style=skillmd]
---
name: repo-skill
description: Repository-specific guidance for working in this codebase. Always consult before making changes.
---

# Working in this repository

## Discover before editing
- Read the project's README, CONTRIBUTING.md, and any AGENTS.md / CLAUDE.md before non-trivial edits.
- Use Glob and Grep to locate relevant code paths instead of guessing file names.
- When you change a public API, search for all callers first.
- Only assert facts you can verify by reading actual files. Do not assume function signatures, class members, or package names without reading the relevant source file first.

## Run tests early and iterate from failures
- Find the test runner from the build manifest (`build.gradle.kts`) before guessing commands.
- Run the relevant test task early — failing test names and assertions are the fastest way to determine what code needs to be written or fixed.
- Run a small, fast subset first (one test file or one module) before running the full suite.
- When a test fails, read the full error output and the failing test source before changing code.

## Make minimal changes
- Match existing style, naming, and abstractions. Don't introduce new patterns unless asked.
- Prefer fixing the root cause over adding compensating logic elsewhere.
- Don't touch unrelated files. Never create new files unless the task explicitly requests them.

## Repository structure

This is a JVM/Kotlin project built with Gradle (Kotlin DSL). Depending on the variant you are working in, the layout may be a single module or a multi-module project:

**Single-module layout:** root contains `build.gradle.kts`, `gradle.properties`, and `src/`.

**Multi-module layout:** root contains `build.gradle.kts` and named subproject directories (e.g. `tracing/`, `eval/`). Each subproject has its own `build.gradle.kts`.

Build and test commands:
- Compile main sources: `./gradlew compileKotlin`
- Compile test sources: `./gradlew compileTestKotlin`
- Run tests: `./gradlew test`
- Full build: `./gradlew build`

For multi-module projects:
- Kotlin Multiplatform modules use `./gradlew :<module>:jvmTest`
- Plain JVM modules use `./gradlew :<module>:test`
- If unsure, run `./gradlew :<module>:tasks` first

A change is not complete until the relevant test task reports BUILD SUCCESSFUL.

## Scoping exploration to the relevant module

Before reading any files, identify which module is relevant to the task by reading the task description and mapping it to a specific subproject directory. Stay within that module during exploration; do not read unrelated modules unless the task explicitly involves them.

- Identify the module first, then explore only its source sets.
- When a symbol appears unresolved in a test file, search for its definition within that module's `src/` tree (checking `commonMain`, `jvmMain`, and any subdirectories) before looking elsewhere.
- Verify that any function or extension you add or reference is exported from the correct source set and package so the test source set can see it.

## Visibility and access conventions

- `internal object` members accessed from other files in the same module must be at least `internal`, not `private`.
- A property setter can have narrower visibility than the getter (e.g. `internal var foo: String = "0"\n    private set`).
- If `compileKotlin` fails with _"Cannot access '…': it is private"_, locate the declaration with `grep -r 'SYMBOL_NAME' --include='*.kt' src/` and widen its visibility modifier.
- For a single-modifier fix, skip broad exploration: grep → edit → `./gradlew compileKotlin`.

## API-change discipline

When adding, removing, or renaming a parameter on any function, method, or interface:
1. Grep the entire source tree for all call sites **and all implementing classes** before editing.
2. Update every call site and every implementor in the same change.
3. Update any README or documentation files that reference the symbol.
4. Verify compilation with `./gradlew compileKotlin` before running tests.

When an interface method is renamed or a new method is added, use `grep` for the interface name (not only the method name) to locate every implementor across the repository. A partial update causes a compilation failure that blocks all tests.

## Kotlin Multiplatform (`expect`/`actual`)

When adding or modifying an API in `commonMain`, provide a corresponding `actual` implementation in **every** target source set (`jvmMain`, `jsMain`, etc.). Forgetting an `actual` declaration causes a compilation error.

Before any rename, search all source sets simultaneously:
```
grep -r OLD_NAME --include="*.kt" src/
```
This surfaces occurrences in `commonMain`, `jvmMain`, `jsMain`, and all test source sets at once. Do not rely on partial searches that might miss a source set.

## Tracing and annotation conventions

- `@KotlinFlowTrace` attribute handlers must be Kotlin `object` singletons (not classes); the JVM implementation calls `.objectInstance` and errors if it is null.
- Call `TracingFlowProcessor.setup()` (or `TracingManager.setup()`) before using any traced method.
- Tracing does not work with `suspend` functions.
- Compile with `-java-parameters` to make parameter names visible at runtime in traces.

## Test dependencies

Before editing `build.gradle.kts`, read all test source imports to identify required artifacts:
```
grep -r '^import' src/test/kotlin/ | sort -u
```
Cross-reference every import against the dependencies already declared in `build.gradle.kts`. Any import whose artifact is not present must be added as a `testImplementation` dependency before compilation.

- `org.junit.jupiter.params.*` requires `junit-jupiter-params` (not included transitively by `kotlin-test-junit5`).
- Testcontainers usage (`@Testcontainers`, `GenericContainer`, `@Container`, `Wait`) requires `org.testcontainers:testcontainers` and `org.testcontainers:junit-jupiter`.
- `com.github.dockerjava` types (`ExposedPort`, `HostConfig`, `Ports`) require `com.github.docker-java:docker-java-api`.
- Match the version of any newly added artifact to what is already resolved: `./gradlew dependencies --configuration testRuntimeClasspath`.

After adding dependencies, verify with `./gradlew compileTestKotlin` before running tests.

## Resolving compilation errors

When `compileKotlin` or `compileTestKotlin` fails with "Unresolved reference":
1. Read the exact unresolved symbol names from the compiler errors.
2. Find every file that imports or references those symbols.
3. Determine whether the fix belongs in production source (most common) or test source.
4. Search for where the symbol should be defined, guided by how it is used at the call sites.
5. Add or correct the definition, then re-run the compile task to confirm the error is gone.
6. Do not consider the change complete until compilation succeeds and tests pass.

## Diagnosing test failures

When a test fails:
1. Read the failing test's assertions first to understand exactly what value, format, or behavior is expected.
2. Follow the call chain from the assertion back into production code — for example, if a test checks a serialized string, read the `@SerialName` annotations and serialization logic; if it checks span attributes, read the attribute keys and how they are set.
3. Form a hypothesis about the root cause before editing. If the root cause is ambiguous, read one more layer of the call chain to reduce uncertainty.
4. After making a change, run the affected test to verify the fix. If it still fails, re-read the assertion and revise the hypothesis rather than making additional speculative edits.
\end{lstlisting}

\newpage{}

\section{GEPA generated SKILL for koog}\label{app:skill}

\lstinputlisting[style=skillmd]{appendix-skills/koog_skill.md}

\newpage{}

\section{GEPA generated SKILL for kotest}\label{app:skill-kotest}

\lstinputlisting[style=skillmd]{appendix-skills/kotest_skill.md}

\newpage{}

\section{GEPA generated SKILL for ktor}\label{app:skill-ktor}

\lstinputlisting[style=skillmd]{appendix-skills/ktor_skill.md}

\newpage{}

\section{SkillOpt generated SKILL for koog}\label{app:skill-skillopt}

\lstinputlisting[style=skillmd]{appendix-skills/koog_skillopt_skill.md}

\newpage{}

\section{A SKILL optimized over per-PR bases}\label{app:skill-outdated}

The document below was produced by the configuration \cref{sec:tasks} rejects: each task placed at its own parent commit, on \ghrepo{JetBrains/tracy}. The repository's drift over the mined window surfaces as the document's opening premise --- two \textit{``repository variants''} the agent is told to distinguish --- and as detailed layouts and APIs that hold at no single commit.

\lstinputlisting[style=skillmd]{appendix-skills/tracy-outdated-skill.md}

\newpage{}

\section{Agent patches for the two open koog issues}\label{app:issue-patches}

The six patches behind \cref{tab:issues}, reviewed anonymized by the maintainer.

\subsection*{Issue \#1275, no SKILL}
\lstinputlisting[style=skillmd]{appendix-patches/patch-koog-1275-dj3m2k.patch}

\subsection*{Issue \#1275, GEPA SKILL}
\lstinputlisting[style=skillmd]{appendix-patches/patch-koog-1275-eh763k.patch}

\subsection*{Issue \#1275, SkillOpt SKILL}
\lstinputlisting[style=skillmd]{appendix-patches/patch-koog-1275-opt2-vs34s2.patch}

\subsection*{Issue \#1354, no SKILL}
\lstinputlisting[style=skillmd]{appendix-patches/patch-koog-1354-eb43i9.patch}

\subsection*{Issue \#1354, GEPA SKILL}
\lstinputlisting[style=skillmd]{appendix-patches/patch-koog-1354-7c72js.patch}

\subsection*{Issue \#1354, SkillOpt SKILL}
\lstinputlisting[style=skillmd]{appendix-patches/patch-koog-1354-opt2-dksd2k.patch}

\end{document}